%% file: main.tex
\documentclass[letterpaper, 10 pt, journal, twoside]{IEEEtran}
\usepackage[linesnumbered,ruled,vlined]{algorithm2e}
\usepackage{algorithmic}
\usepackage{amsmath,amsfonts,amssymb}

\usepackage{tikz}
\usepackage{graphicx}
\graphicspath{{figures/}}
\usepackage{epsfig} % for postscript graphics files
\usepackage{times}
\usepackage{siunitx}
\usepackage{hyperref}
\usepackage[capitalise]{cleveref}
\usepackage[inline]{enumitem}
\usepackage[normalem]{ulem}

\Crefname{algorithm}{Alg.}{Algs.}
\Crefname{section}{Sec.}{Secs.}
\Crefname{equation}{Eq.}{Eqs.}

\usepackage{authblk}
\usepackage{booktabs}
\usepackage{caption}
\usepackage[
    sorting=none,
    giveninits=true,
    maxbibnames=9,
    maxcitenames=2,backend=biber
    ]{biblatex}
\title{Long-Term Human Motion Prediction Using Spatio-Temporal Maps of Dynamics}

\author{
Yufei Zhu$^{1}$, Andrey Rudenko$^{2}$, Tomasz P. Kucner$^{3}$, Achim J. Lilienthal$^{1,2}$, Martin Magnusson$^{1}$%
\thanks{Manuscript received: May 29, 2025; Revised August 31, 2025; Accepted September 23, 2025. This paper was recommended for publication by Editor Markus Vincze upon evaluation of the Associate Editor and Reviewers' comments. This work received funding from the European Union’s Horizon 2020 research and innovation programme under grant agreement No 101017274 (DARKO) and 101070596 (euRobin). \textit{(Corresponding author: Yufei Zhu.)}}%
\thanks{$^{1}$Yufei Zhu, Achim J. Lilienthal, and Martin Magnusson are with the Centre for Applied Autonomous Sensor Systems (AASS), \"Orebro University, Sweden 
{\tt\footnotesize \{yufei.zhu, achim.lilienthal, martin.magnusson\}@oru.se}}%
\thanks{$^{2}$Andrey Rudenko and Achim J. Lilienthal are with Technical University of Munich, Germany
{\tt\footnotesize \{andrey.rudenko, achim.j.lilienthal\}@tum.de}}%
\thanks{$^{3}$Tomasz P. Kucner is with department of electrical engineering and automation, school of electrical engineering, Aalto University, Finland
{\tt\footnotesize tomasz.kucner@aalto.fi}}%
\thanks{Digital Object Identifier (DOI): see top of this page.}
}

\newif\ifkeepremark
\newcommand{\add}[1]{
    \ifkeepremark
    \textcolor{teal}{#1}
    \else
    #1
    \fi
}

\newcommand{\replace}[2]{
    \ifkeepremark
    \textcolor{red!60!black}{\sout{#1}}\textcolor{teal}{#2}
    \else
    #2
    \fi
}

\newcommand{\kstef}{k_\mathrm{stef}}

\keepremarkfalse
\normalsize

\begin{document}

\maketitle
%\thispagestyle{empty}
%\pagestyle{empty}

\input{revision-content/commands}

\input{revision-content/abstract}

%%%%%%%%%%%%%%%%%%%%%%%%%%%%%%%%%%%%%%%%%%%%%%%%%%%%%%%%%%%%%%%%%%%%%%%%%%%%%%%%

\input{revision-content/introduction}

\input{revision-content/related_work}

\input{revision-content/method}

\input{revision-content/experiments}

\input{revision-content/results}

\input{revision-content/conclusions}

% \section*{Acknowledgments}
% This should be a simple paragraph before the References to thank those individuals and institutions who have supported your work on this article.

% {\appendix[Proof of the Zonklar Equations]
% Use $\backslash${\tt{appendix}} if you have a single appendix:
% Do not use $\backslash${\tt{section}} anymore after $\backslash${\tt{appendix}}, only $\backslash${\tt{section*}}.
% If you have multiple appendixes use $\backslash${\tt{appendices}} then use $\backslash${\tt{section}} to start each appendix.
% You must declare a $\backslash${\tt{section}} before using any $\backslash${\tt{subsection}} or using $\backslash${\tt{label}} ($\backslash${\tt{appendices}} by itself
%  starts a section numbered zero.)}


\printbibliography

%\vfill

\end{document}

%% file: revision-content/commands.tex
%
% Common abbreviations
%
\newcommand{\ie}{i.\,e.}
\newcommand{\eg}{e.\,g.}
\newcommand{\etc}{etc.}

%
% Environments
%
\newtheorem{definition}{Definition}

%
% Editing commands
%
\newcommand{\citn}{{\color{red}[Citation needed]}} % citation to be addedd
\newcommand{\refn}{{\color{red}[Reference needed]}} % reference to a forward section
\newcommand{\reflect}[2][]{{\color{blue}[\ifthenelse{\equal{#1}{}}{}{(#1) }#2]}} % part requires reflection - optional argument to specify who added the comment
\newcommand{\tbd}{{\color{orange}TBD}} % to be done
\newcommand{\com}[2][]{{\color{green}[\ifthenelse{\equal{#1}{}}{}{(#1) }#2]}} % comment - optional argument to specify who added the comment
\newcommand{\todo}[1]{{\color{red}TODO:#1}}
\newcommand{\wip}{{\color{orange}WIP}}
\newcommand{\rev}{{\color{yellow}Revision}}
\newcommand{\done}{{\color{green}Done}}
\newcommand{\att}[2]{{\color{orange}{\textbf{[(@#1) #2}]}}} %"att" for targeted todos

%% file: revision-content/abstract.tex
\begin{abstract}

%\todo{Write one statement about the general problem}
%\todo{What is this paper is about}
%\todo{Describe how it solves the problem}
%\todo{Emphasize what is new, or better}
%\todo{Mention the evidence indicating the advantages of the proposed approach}

Long-term human motion prediction (LHMP) is important for the safe and efficient operation of autonomous robots and vehicles in environments shared with humans. Accurate predictions are important for applications including motion planning, tracking, human-robot interaction, and safety monitoring. In this paper, we exploit Maps of Dynamics (MoDs), which encode spatial or spatio-temporal motion patterns as environment features, to achieve LHMP for horizons of up to 60 seconds. We propose an MoD-informed LHMP framework that supports various types of MoDs and includes a ranking method to output the most likely predicted trajectory, improving practical utility in robotics. Further, a time-conditioned MoD is introduced to capture motion patterns that vary across different times of day. We evaluate MoD-LHMP instantiated with three types of MoDs. Experiments on two real-world datasets show that MoD-informed method outperforms learning-based ones, with up to 50\% improvement in average displacement error, and the time-conditioned variant achieves the highest accuracy overall. Project code is available at \url{https://github.com/test-bai-cpu/LHMP-with-MoDs.git}
\end{abstract}

\begin{IEEEkeywords}
Human Detection and Tracking, Human and Humanoid Motion Analysis and Synthesis, Probability and Statistical Methods, Human-Aware Motion Planning
\end{IEEEkeywords}

%Also, we introduce a time-conditioned variant of the previous approach CLiFF-LHMP, evaluated alongside LHMP with the STeF-map representation and deep learning based baselines. 

%Evaluations on two real-world datasets show that the MoD-informed LHMP approach outperforms deep learning-based methods. Moreover, the time-conditioned variant achieves higher prediction accuracy.

%% file: revision-content/introduction.tex
\section{Introduction}
\IEEEPARstart{E}{nsuring} safe and efficient operation of robots in complex and dynamic environments, particularly in the presence of humans, is critical for deploying robotic systems to assist with a wide range of real-world tasks \cite{triebel2016spencer, molina2023iliad}. A key element in achieving this goal is long-term human motion prediction, i.e., anticipating the trajectories of individuals over extended periods. Accurate long-term prediction of future trajectories is a fundamental requirement for various applications, including optimized motion planning, refined tracking, advanced automated driving, improved human-robot interaction, and enhanced intelligent safety monitoring and surveillance. 

Human motion is complex, influenced by various factors. These include not only an individual's intrinsic intent and dynamics but also external influences such as social conventions and environmental cues~\cite{helbing1995social}. Predicting trajectories over extended periods, such as 20 seconds or more, requires careful consideration of the impact of large-scale environments on human behavior. While short-term predictions can often rely on current state and immediate interactions, long-term predictions demand more comprehensive modeling to capture how the environment influences and guides human movement.

An effective approach to address long-term human motion prediction (LHMP) is through the use of maps of dynamics (MoDs), which encode spatial or spatio-temporal motion patterns as a feature of the environment. Prior work, CLiFF-LHMP~\cite{zhu2023clifflhmp}, exploits the CLiFF-map representation~\cite{kucner2017enabling}, which is a specific type of MoD that stores a multi-modal, continuous joint distribution of speed and orientation for each discrete map location, to predict human trajectories over long-term horizons. In this work, we extend CLiFF-LHMP to a general MoD-informed LHMP framework, named \emph{MoD-LHMP}, which can be applied with various types of MoDs. By using MoDs, motion prediction can utilize previously observed motion patterns. We also introduce a \emph{ranking} method that enables the prediction of the most likely trajectory output, enhancing its practical utility for robotic applications. 

We instantiate MoD-LHMP with multiple types of MoDs, including: (1) the original CLiFF-map; 
(2) a Time-Conditioned CLiFF-map, which we introduce in this work; 
and (3) STeF-map~\cite{molina22exploration}, a spatio-temporal MoD designed to capture periodic motion patterns. Given that human movement in the same environment varies over time, in this work, we address the problem of capturing spatio-temporal motion patterns and use them for LHMP. We present the Time-Conditioned CLiFF-map, which adapts to varying motion patterns at different times of day.
The methods are also compared with Trajectron++~\cite{salzmann20}, LSTM-based human motion prediction methods~\cite{alahi2016social},\add{a diffusion-based model~\cite{gu2022mid} and a transformer-based model~\cite{shi2023tutr}}.
The evaluation uses two real-world datasets: the ATC~\cite{brscic2013person} and the Edinburgh dataset~\cite{majecka2009statistical},\add{both capturing open indoor environments.} Both datasets span multiple days and exhibit variations in human motion patterns throughout the day. Through this comparative study, we aim to evaluate the performance of spatio-temporal MoDs in the LHMP task.

In summary, we make the following contributions:
\begin{itemize}
\item \add{We extend CLiFF-LHMP by introducing a ranking method, enabling most-likely output predicted trajectory, improving its practical applicability in robotics.}
\item \add{We demonstrate how the framework can be instantiated with a range of different map representations, thus providing a general MoD-LHMP framework.}
\item We introduce Time-Conditioned CLiFF-map, a temporal variant of CLiFF-map, to improve prediction accuracy.
\item \add{We analyze how different instantiations of MoD-LHMP perform on two real-world datasets, compared with learning-based baselines.}

\end{itemize}

%% file: revision-content/related_work.tex
\vspace{-2mm}
\section{Related work} \label{section-relatedwork}
%TODO:
%3. To do: add references from ICRA 2023. How can we relate to the work of the 
% density maps of Ge et al, 
% the road networks of Schmidt et al. Exploring Navigation Maps for Learning-Based Motion Prediction ? 
% Ding et al, 
% Wang et al?

Human motion prediction has been studied extensively in recent years. With different prediction horizons, the human motion prediction problem can be short-term (1--2~s), long-term (up to 20~s) \cite{rudenko2020human}, and extended long-term (which we define as over 20~s). In this work, we focus on extended LHMP.

Based on the underlying principle for the motion model, motion prediction approaches can be categorized into planning-, pattern- and physics-based approaches~\cite{rudenko2020human}. In planning-based methods, prior knowledge of potential goals in the environment can be used. \textcite{ikeda2013modeling} use the concept of sub-goals, i.e., way-points that pedestrians tend to\replace{pass}{move toward}before reaching the final destination. After retrieving the position of the sub-goals, this method models the long-term behavior of pedestrians through transition probabilities between sub-goals. %Information about interaction with objects in the environment can also be used to predict the navigation goal of a moving human.
\textcite{bruckschen2019human} propose an approach that uses knowledge about typical human-object interaction sequences. This method learns a transition model of subsequent object interaction and utilizes it to infer the navigation goal using a recursive Bayes' filter. 
\textcite{Gaorlo2024SceneGraph} predicts extended long-term (up to \SI{60}{\second}) human trajectories using large language models to reason about human interactions with the scene, represented as a 3D dynamic scene graph.

Another main type of approaches is clustering-based methods, which belong to the category of pattern-based approaches. These approaches cluster observed trajectories to create a set of long-term motion patterns. \textcite{bennewitz2005learning} clusters full trajectories into motion patterns and uses hidden Markov models derived from the learned patterns for prediction. \textcite{chen2008pedestrian} propose a dynamic clustering method that can learn the motion pattern which change over time. In the work by \textcite{bera2016glmp}, global and local motion patterns are learned using Bayesian inference in real-time. One shortcoming of clustering-based methods is their reliance on complete trajectories as input. In practice, especially when using a robot's on-board sensors in cluttered environments, long trajectories are hard to be observed from start to finish. And it is difficult to cluster shorter, incomplete trajectories in a meaningful way.

Clustering-based methods are non-sequential and directly model the distribution over full trajectories. In contrast to that, other pattern-based approaches are sequential and assume that human motion can be described with causally conditional models over time. %Sequential predictions are generated by learning local motion patterns. 
In the works of \cite{thompson2009probabilistic, ballan2016knowledge}, models of local motion transition patterns are proposed. There are also approaches for pedestrian crowd prediction. \textcite{Kiss2022GP} use a constrained Gaussian process to give a smooth and continuous representation of the crowd dynamics into the future.

In addition to above approaches, physics-based methods predict human motion using kinematic models, typically without modeling underlying forces. 
One popular example is the constant velocity model (CVM), which is the simplest approach to predict human motion. \textcite{scholler2019simpler} showed that CVM can outperform even state-of-the-art neural models at a 4.8~s prediction horizon, but CVM is not reliable for long-term prediction as it ignores environment information. 

%One approach to predict long-term human motion is to account for various semantic attributes of the static environment. In this paper, 
Our approach to LHMP is connected to the field of maps of dynamics (MoDs). MoDs are maps that explicitly encode changes (e.g., motion). By building spatial and spatio-temporal models, MoDs can capture the patterns followed by dynamic objects (such as humans) in the environment, making them effective for human motion prediction~\cite{zhu2023clifflhmp, zhu2024lace}.

There are several approaches for creating maps of human motion.\add{Occupancy-based methods focus on mapping human dynamics on occupancy grid maps, modeling motion as shifts in occupancy~\cite{wang2016building}. Trajectory-based methods extract human trajectories and group them into clusters, with each cluster representing a typical path through the environment~\cite{bennewitz2005learning}. These approaches suffer from noisy or incomplete trajectories. To address this, \textcite{chen2016augmented} formulate trajectory modeling as a dictionary learning problem and use augmented semi-nonnegative sparse coding to find local motion patterns characterized by partial trajectory segments.
%Velocity-based methods focuses on mapping local velocity observations into flow fields, without requiring complete trajectories. 
}

%There are several approaches for creating maps of human motion. As in \cite{wang2016building}, one type of MoD method focuses on mapping human dynamics in occupancy grid maps. Another type of MoDs is mapping human trajectories, which is used in the aforementioned clustering-based human motion prediction approach \cite{bennewitz2005learning}. 

MoDs can also be based on velocity observations. With velocity mapping, human dynamics can be modeled through flow models. 
\textcite{kucner2017enabling} presented a probabilistic framework for mapping velocity observations, which is named Circular-Linear Flow Field map (CLiFF-map). CLiFF-map can address multimodality in the data and this characteristic can be utilized for LHMP. Also, in contrast to MoDs that map trajectories, CLiFF-map allows building maps of motion patterns from incomplete or spatially sparse data~\cite{almeida2024performance}. While originally constructed offline, recent work has proposed online CLiFF-maps that update models with new observations~\cite{zhu2025cliffonline}.

When building flow models, temporal information can also be incorporated. \textcite{molina22exploration} apply the Frequency Map Enhancement (FreMEn~\cite{krajnik2017fremen}), which is a model describing spatio-temporal dynamics in the frequency domain, to build a time-dependent probabilistic map to model periodic changes in people flow, called STeF-map. The motion orientations in STeF-map are discretized. Another method of incorporating temporal information is proposed by \textcite{zhi2019spatiotemporal}. Their approach uses a long-short term memory network to provide a multimodal probability distribution of movement directions of a typical object in the environment over time.

%In this work, we focus on extended long-term human motion predictions up to 60~s. Our approach exploits MoDs to predict human motions. The approach can be applied with several MoDs that represent velocities.

%% file: revision-content/method.tex
\vspace{-2mm}
\section{Method} \label{section-method}
% In this section, firstly the problem is formally presented in \cref{problem-formulation}. Then we present a general approach for single-agent long-term human motion prediction using Maps of Dynamics (\cref{mod-lhmp}), named MoD-LHMP. In MoD-LHMP, multiple MoDs that represent velocity can be used. The MoDs demonstrated in this work (\cref{examples-mod-lhmp}) are CLiFF-map, Time-Conditioned CLiFF-map and STeF-map. For each MoD, we present the method of using it for human motion prediction. Finally, in \cref{rank}, we describe the ranking method for trajectories predicted with MoD-LHMP.

\subsection{Problem Formulation} \label{problem-formulation}

%\begin{algorithm}[t]
%\small
%    \KwIn{$\mathcal{H}$, $x_{t_0}$, $y_{t_0}$, $\Xi$}
%    \KwOut{$\mathcal{T}$}
%	%\SetAlgoLined
%	$\mathcal{T} = \{\}$ \
	
%	$\rho_\mathrm{obs}, \theta_\mathrm{obs} \leftarrow $  getObservedVelocity($\mathcal{H}$) \
	
%	$s_{t_0} = (x_{t_0},y_{t_0},\rho_{\mathrm{obs}},\theta_{\mathrm{obs}})$ \
	
%	\For { $t= t_{0}+1$, ..., $t_{0}+T_p$}{
	
%	    $x_t, y_t \leftarrow $ getNewPosition($s_{t-1}$) \
	    
%		% $\xi \leftarrow $ selectSWGMM($x_t, y_t$)\
	    
%		$\rho_s$, $\theta_s$ $ \leftarrow $ sampleVelocityFromMoD($x_t, y_t, \Xi$)\
	    
%		$\rho_t$, $\theta_t$ $ \leftarrow $ predictVelocity($\rho_s$, $\theta_{s}$, $\rho_{t\textendash1}$, $\theta_{t\textendash1}$)\
		
%		$s_t \leftarrow (x_t, y_t, \rho_{t}, \theta_t)$
		
%		$\mathcal{T} \leftarrow \mathcal{T} \cup s_t$ \
%    }
 %   \Return $\mathcal{T}$ \
%\caption{MoD-LHMP}
%\label{alg:LHMPAlgo}
%\end{algorithm}

\begin{algorithm}[t]
\small
    \KwIn{$\mathcal{H}$, $x_{t_0}$, $y_{t_0}$, $\Xi$}
    \KwOut{$\mathcal{T}$, \textcolor{red}{$p$}}
	%\SetAlgoLined
	$\mathcal{T} = \{\}$ \
	
	$\rho_\mathrm{obs}, \theta_\mathrm{obs} \leftarrow $  getObservedVelocity($\mathcal{H}$) \
	
	$s_{t_0} = (x_{t_0},y_{t_0},\rho_{\mathrm{obs}},\theta_{\mathrm{obs}})$ \

        \textcolor{red}{$p = 1$}
	
	\For { $t= t_{0}+1$, ..., $t_{0}+T_p$}{
	
	    $x_t, y_t \leftarrow $ getNewPosition($s_{t\textendash1}$) \
	    
		% $\xi \leftarrow $ selectSWGMM($x_t, y_t$)\
	    
		$\rho_s$, $\theta_s$, \textcolor{red}{$p_t$} $ \leftarrow $ sampleVelocityFromMoD($x_t, y_t, \Xi$)\
	    
		$\rho_t$, $\theta_t$ $ \leftarrow $ predictVelocity($\rho_s$, $\theta_{s}$, $\rho_{t\textendash1}$, $\theta_{t\textendash1}$)\
		
		$s_t \leftarrow (x_t, y_t, \rho_{t}, \theta_t)$

            \textcolor{red}{$p \leftarrow p * p_t$}
		
		$\mathcal{T} \leftarrow \mathcal{T} \cup s_t$ \
    }
    \Return $\mathcal{T}$, \textcolor{red}{$p$} \

 %\caption{Motion prediction using CVM and MoD}
 \caption{MoD-LHMP}
\label{alg:LHMPAlgo}
\end{algorithm}

\begin{algorithm}[t]

\small
    \KwIn{$x$, $y$, $\Xi$}
    \KwOut{$\rho_s$, $\theta_s$}
	$\Xi_{\mathrm{near}} \leftarrow $ getNearSWGMMs($x, y, \Xi$) \
	
	$\xi \leftarrow $ selectSWGMM($\Xi_{\mathrm{near}}$) \
	
	$\rho_s$, $\theta_s$, \textcolor{red}{$p$} $\leftarrow $ sampleVelociyFromSWGMM($\xi$) \
    
    \Return $\rho_s$, $\theta_s$, \textcolor{red}{$p$}
 \caption{sampleVelocityFromCLiFF($x, y, \Xi$)}
\label{alg:samplecliff}
\end{algorithm}

\begin{algorithm}[t]
\small
    \KwIn{$x$, $y$, $\Xi$, $\rho_{t\textendash1}$}
    \KwOut{$\rho_s$, $\theta_s$}
	$\Xi_{\mathrm{near}} \leftarrow $ getNearCell($x, y, \Xi$) \
    
	$\theta_s$, \textcolor{red}{$p$} $\leftarrow $ sampleDirectionFromCell($\xi$) \

        $\rho_s \leftarrow \rho_{t-1}$ \
    
    \Return $\rho_s$, $\theta_s$, \textcolor{red}{$p$}
 \caption{\add{sampleVelocityFromSTeF($x, y, \Xi, \rho_{t\textendash1}$)}}
\label{alg:samplestef}
\end{algorithm}

Predicting a person's future trajectory is framed as using the past trajectory to estimate a sequence of future states. The observation length is $O_s \in \mathbb{R}^+$ seconds, equivalent to an integer $O_p > 0$ time steps. With the current time-step denoted as the integer $t_0 \geq 0$, the sequence of observed states is $\mathcal{H} = \langle s_{t_{0} - 1},..., s_{t_{0} - O_p} \rangle$, where $s_t$ is the state of a person at time-step $t$. A state is represented by 2D Cartesian coordinates $(x,y)$, speed $\rho$ and orientation $\theta$: $s = (x,y,\rho, \theta)$.

\add
{To predict the future trajectory, we estimate the pedestrian's velocity $(\rho_{\mathrm{obs}},\theta_{\mathrm{obs}})$ at current state from the observed sequence $\mathcal{H}$. Following the ATLAS benchmark~\cite{rudenko2022atlas}, we compute a weighted average of recent velocities using a zero-mean Gaussian kernel with $\sigma = 1.5$, which assign higher weights to more recent observations, such that $\rho_{\mathrm{obs}} = \sum_{t=1}^{O_p}v_{t_0 - t}g(t)$ and $\theta_{\mathrm{obs}} = \sum_{t=1}^{O_p}\theta_{t_0 - t}g(t)$, where $g(t) = (\sigma\sqrt{2\pi}e^{\frac{1}{2}(\frac{t}{\sigma})^2})^{-1}$. The current state at $t_0$ is then represented as $s_{t_0} = (x_{t_0},y_{t_0},\rho_{\mathrm{obs}},\theta_{\mathrm{obs}})$.}
%From the observed sequence $\mathcal{H}$, we derive the observed speed $\rho_{\mathrm{obs}}$ and orientation $\theta_{\mathrm{obs}}$ at time-step $t_0$. Then the current state becomes $s_{t_0} = (x_{t_0},y_{t_0},\rho_{\mathrm{obs}},\theta_{\mathrm{obs}})$. The values of $\rho_{\mathrm{obs}}$ and  $\theta_{\mathrm{obs}}$ are calculated as a weighted sum of the finite differences in the observed states, as in the popular ATLAS benchmark \cite{rudenko2022atlas}. 

From the current state $s_{t_0}$, the goal is to estimate a sequence of future states
%. Similar to past states, future states are predicted within horizon
up to $T_s \in \mathbb{R}^+$ \SI{}{\second}. $T_s$ is equivalent to $T_p$ prediction time steps assuming the constant time interval $\Delta t$. %between two predictions.
Thus, the prediction horizon is $T_s = T_p \Delta t$. The future sequence is then denoted as $\mathcal{T} = \langle s_{t_0+1}, s_{t_0+2},...,s_{t_0+T_p} \rangle$.
\vspace{-1mm}
\subsection{MoD-LHMP} \label{mod-lhmp}
CLiFF-LHMP \cite{zhu2023clifflhmp} was the first method to exploit an MoD for long-term human motion prediction. Compared with CLiFF-LHMP which uses a specific type of MoD, CLiFF-map, in this work we extend it to \emph{MoD-LHMP} that can be used with all types of MoDs that represent velocities. MoD-LHMP predicts stochastic trajectories by sampling a velocity from MoDs to guide a velocity filtering model.

The algorithm of MoD-LHMP is presented in \cref{alg:LHMPAlgo}.\add{The basic algorithm is shown, with an extended version highlighted in red that ranks the predicted trajectories includes additional updates, which will be introduced later in \cref{rank}.} In \cref{alg:LHMPAlgo}, with the input of an MoD $\Xi$, past states $\mathcal{H}$ and current location $(x_{t_0}$, $y_{t_0})$ of a person, the algorithm predicts a sequence of future states. To estimate $\mathcal{T}$, for each prediction time step, a velocity $(\rho_s, \theta_s)$ is sampled from the MoD at the current position ($x_t$, $y_t$) to bias the prediction with the learned motion patterns represented by the MoD. The main steps for each prediction iteration are shown in lines 5--9 of \cref{alg:LHMPAlgo}.

In each iteration, the new position of prediction time step $t$ (line 6 of \cref{alg:LHMPAlgo}) is updated from the previous state:
\begin{equation}
\begin{gathered}
    x_{t} = x_{t-1} + \rho_{t-1}\cos{\theta_{t-1}} \Delta t,\\
    y_{t} = y_{t-1} + \rho_{t-1}\sin{\theta_{t-1}} \Delta t,\\
\end{gathered}
\end{equation}

\noindent Then we estimate the new speed and orientation using a biased version of the CVM.
To estimate velocity at $t$, we sample $\rho_s, \theta_s$ from MoD at location $(x_t, y_t)$ in the function \texttt{sampleVelocityFromMoD()} (line 7 of \cref{alg:LHMPAlgo}). This function is the only part of \cref{alg:LHMPAlgo} that depends on the choice of MoD. In principle, any MoD that allows sampling velocities could be used. The instantiations for CLiFF-map and STeF-maps are shown in \cref{alg:samplecliff} and  \cref{alg:samplestef}, respectively.

In line 8, we predict velocity ($\rho_t$, $\theta_t$) by biasing the last step velocity with the sampled one ($\rho_s$, $\theta_s$) as:
\begin{equation}
\begin{gathered}
\rho_t = \rho_{t-1} + (\rho_s - \rho_{t-1}) \cdot K(\rho_{s} - \rho_{t-1}), \\
\theta_t = \theta_{t-1} + (\theta_s - \theta_{t-1}) \cdot K(\theta_{s} - \theta_{t-1}), \\
\end{gathered}
\end{equation}

\noindent where $K(\cdot)$ is a kernel function that defines the degree of impact of the MoD. We use a Gaussian kernel with a parameter $\beta$ that represents the width $K(x) =  e ^ {-\beta \left\Vert x \right\Vert ^ 2}$.

With kernel $K$, the MoD term is scaled by the difference between the velocity sampled from the MoD and the current velocity according to the CVM. The sampled velocity is given less weight if it deviates more from the current velocity. 
This mechanism accounts for outlier trajectories that do not align with the motion patterns encoded in the MoD. A larger $\beta$ value makes the method behave more like a CVM, and a smaller $\beta$ makes it more closely follow the MoD.

At the end of each iteration, $s_t$ is added to the predicted trajectory $\mathcal{T}$ and updated $t$ for the next iteration (line 11 of \cref{alg:LHMPAlgo}). After iterating for $T_p$ times, there is a sequence $\mathcal{T}$ of future states that represents the predicted trajectory.

\vspace{-1mm}
\subsection{Examples of MoD-LHMP}  \label{examples-mod-lhmp}
%In MoD-LHMP, multiple MoDs that represent velocity can be used. Besides the previous CLiFF-map (\cref{method-cliff}), a new variation of CLiFF-map is presented to be used for LHMP, which also considers temporal dimension compared with the original spatial-dependent CLiFF-map, named Time-Conditioned CLiFF-map, presented in \cref{method-tcliff}. With a Time-Conditioned CLiFF-map, human motion patterns that vary over different times in a day can be captured.
In MoD-LHMP, different MoDs can be used to represent velocity. In addition to the original CLiFF-map (\cref{method-cliff}), we introduce a Time-Conditioned CLiFF-map (\cref{method-tcliff}), which extends the spatially dependent CLiFF-map by incorporating the temporal dimension. This allows human motion patterns that vary across different times of day to be captured.
In \cref{method-stef}, MoD-LHMP is instantiated with STeF-map, which builds a spatio-temporal model of human motion using the frequency spectrum of human activities.

\subsubsection{Circular-Linear Flow Field Map (CLiFF-map)}  \label{method-cliff}
CLiFF-map represents motion patterns using multimodal statistics to represent speed and orientation jointly~\cite{kucner2017enabling}.\add{It associates to an arbitrary set of discrete locations a set of Semi-Wrapped Gaussian Mixture Model (SWGMM)~\cite{Roy16SWGMM}}to capture the dependency between the speed and orientation, representing motion patterns based on local observations. 
The sampling velocity function 
%\texttt{sampleVelocityFromMoD()} (line 7 of \cref{alg:LHMPAlgo}) 
for CLiFF-map is illustrated in \cref{alg:samplecliff}. To sample a direction at location $(x,y)$, 
from $\Xi$, we firstly get the SWGMMs $\Xi_{\mathrm{near}}$ whose distances to $(x,y)$ are less than the sampling radius $r_s$. Each SWGMM location in a CLiFF-map is associated with a motion intensity ratio,\add{defined as the ratio of the time during which motion was observed at that location to the total observation time. This ratio provides an estimate of how frequently motion occurs.}The SWGMM $\xi$ with highest motion intensity ratio is selected from $\Xi_{\mathrm{near}}$. The speed and orientation are sampled from the selected SWGMM and returned for motion prediction.
%In the CLiFF-map, each location is associated with a Semi-Wrapped Gaussian Mixture Model (SWGMM)~\cite{Roy16SWGMM} 

\subsubsection{Time-Conditioned CLiFF-map}  \label{method-tcliff}
In previous work, CLiFF-maps have been trained using cumulative data, integrating data regardless of when they occurred. This approach does not account for variations in motion patterns over time. To address this, we introduce Time-Conditioned CLiFF-maps, which represent motion patterns at different times of day and more accurately capture temporal variations in human flow.

\begin{figure*}[t]
\centering
\includegraphics[clip,trim= 20mm 3mm 48mm 6mm,height=32mm]{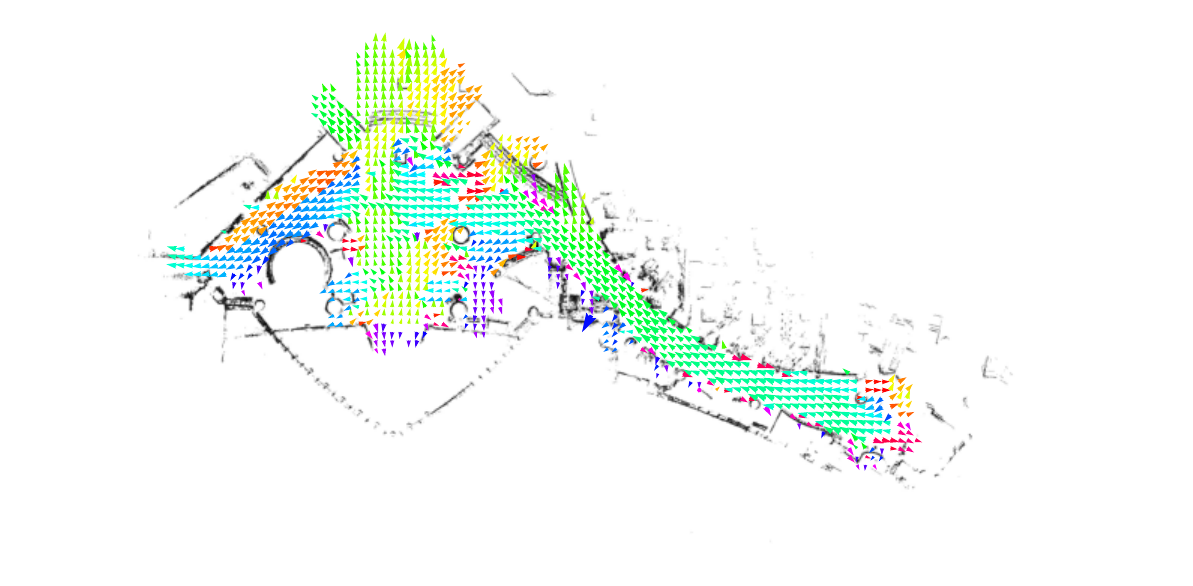}%
\includegraphics[clip,trim= 20mm 3mm 48mm 6mm,height=32mm]{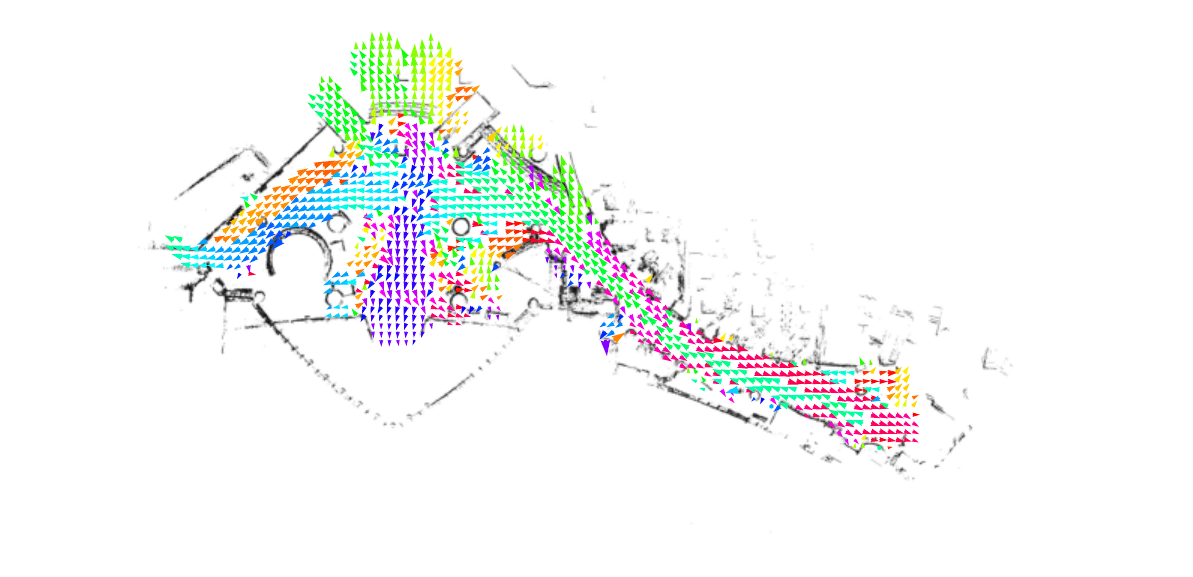}%
\includegraphics[clip,trim= 20mm 3mm 48mm 6mm,height=32mm]{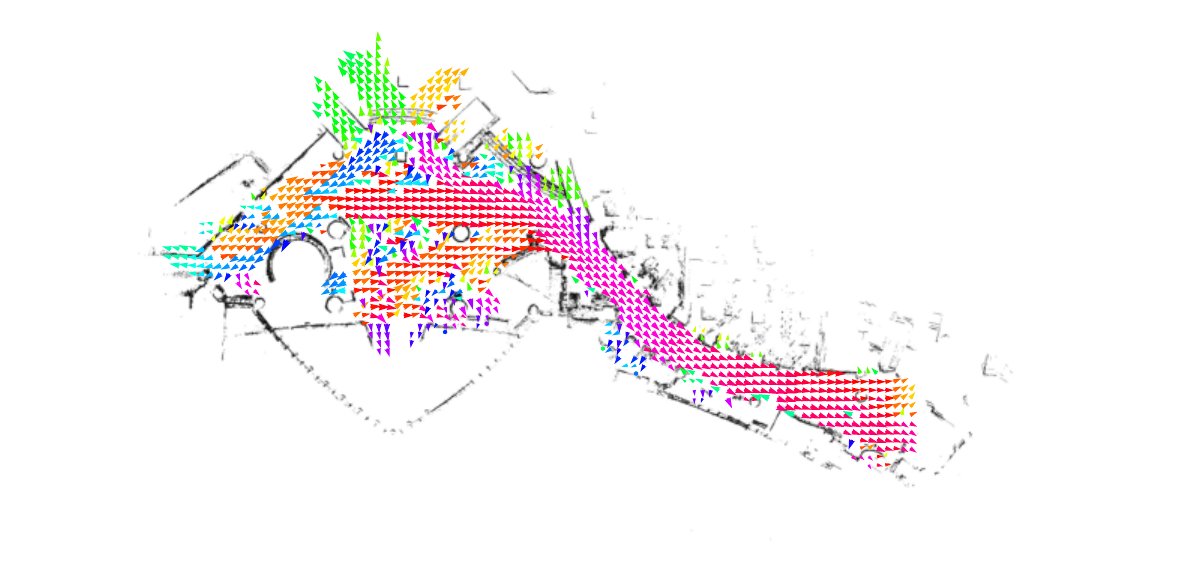}%
\includegraphics[clip,trim= 175mm 0mm 0mm 6mm,height=32mm]{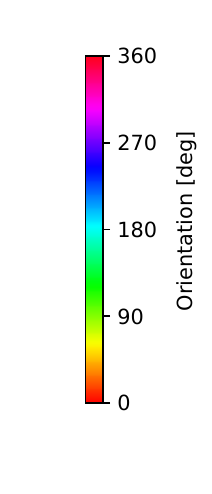}
\vspace*{-6mm}
\caption{Time-conditioned CLiFF-map in the ATC dataset, for 10:00 (\textbf{left}), 14:00 (\textbf{middle}) and 18:00 (\textbf{right}), showing changes of motion patterns throughout the day represented by CLiFF-map. At each location, the colored arrow shows the mean of the Gaussian component with maximum weight,\add{where the arrow color encodes orientation and the arrow length encodes speed.}}
\label{fig:cliff-atc-time}
\vspace*{-5mm}
\end{figure*}

\begin{figure}
\centering
\includegraphics[width=0.8\linewidth]{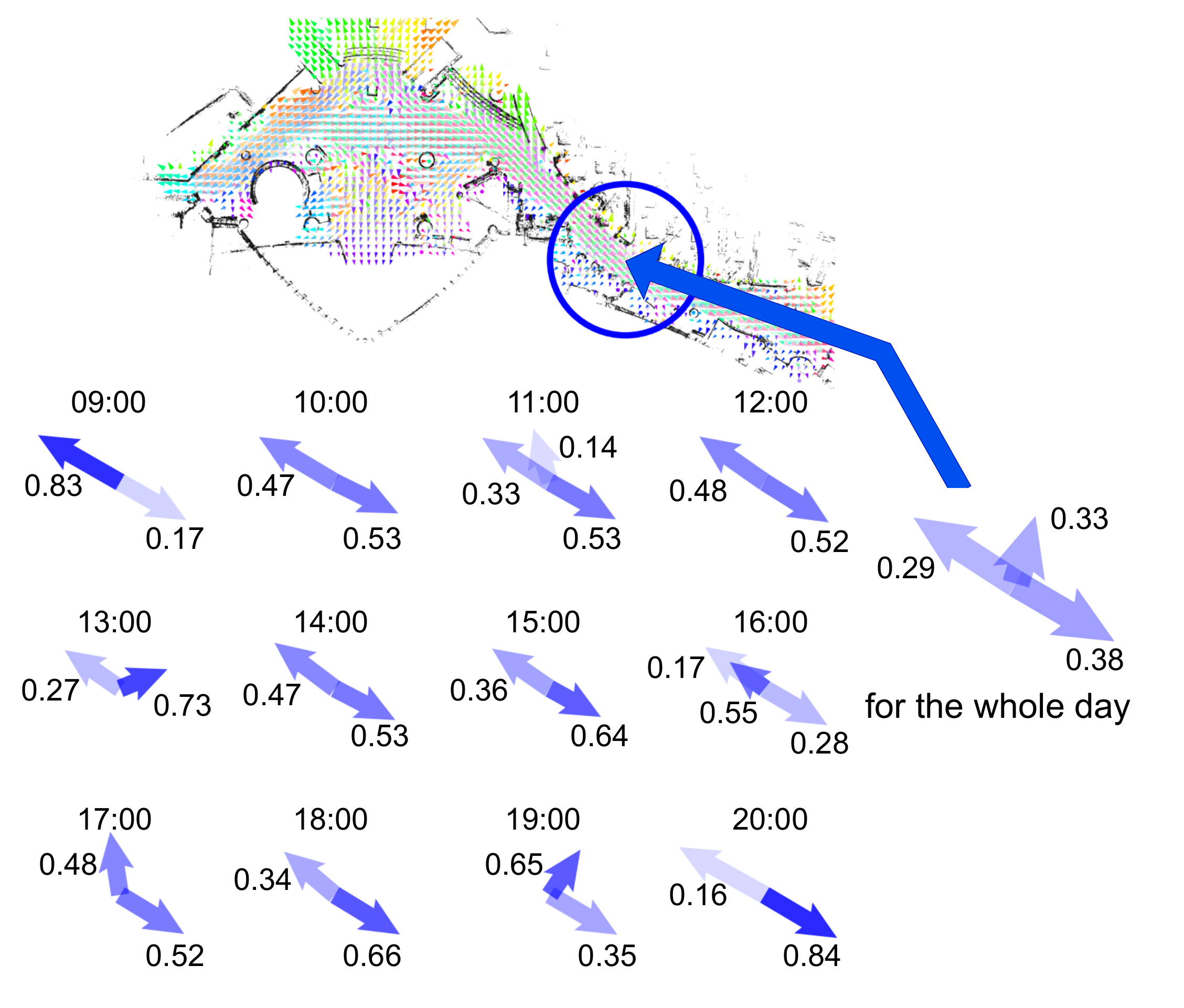}
\vspace*{-2mm}
\caption{A focused view of CLiFF-maps at one location in the east corridor of the ATC dataset. For each hour between 9:00 to 21:00, Time-Conditioned CLiFF-maps of the example location are shown, together with the general CLiFF-map of the whole day at the same location. Arrows show the mean of each component in SWGMM, jointly representing speed and orientation.\add{Arrow length encodes speed, while arrow transparency reflects the component weight (lighter arrows correspond to smaller weights).}}
\label{fig:cliff-atc-time-one-location}
\vspace*{-5mm}
\end{figure}

To build Time-Conditioned CLiFF-maps, a day is divided into $n$ time intervals. For each interval, a separate CLiFF-map is trained using the trajectories that occur  during that time. Consequently, for a single day, $n$ Time-Conditioned CLiFF-maps are generated, one for each time interval. \cref{fig:cliff-atc-time} shows the CLiFF-map of 10:00, 14:00 and 18:00 on the first day of the\add{ATC dataset~\cite{brscic2013person}. Implementation details are provided later in \cref{section-experiments}.}These maps visually demonstrate the variations in human motion patterns throughout the day. For a more focused view, as an example, \cref{fig:cliff-atc-time-one-location} shows the CLiFF-map at a specific location in the east corridor of the ATC environment, showing how motion patterns change hourly. The Time-Conditioned CLiFF-maps provide a more accurate representation of human motion compared to the general CLiFF-map.

\begin{figure*}[t]
\centering
\includegraphics[clip,trim= 20mm 3mm 70mm 6mm,height=32mm]{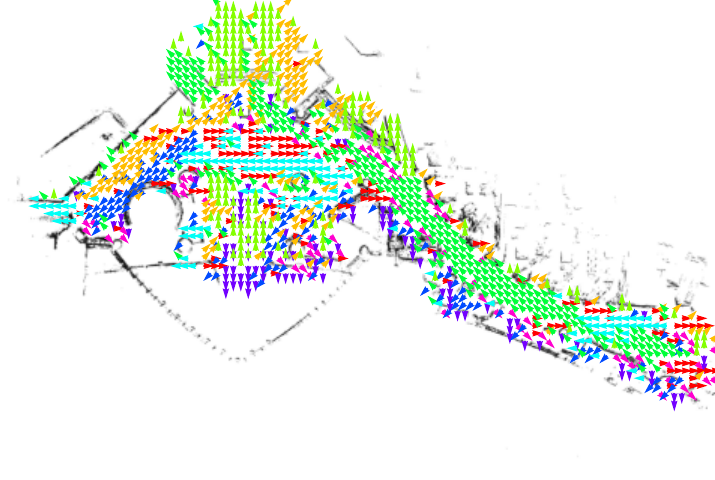}%
\includegraphics[clip,trim= 20mm 3mm 70mm 6mm,height=32mm]{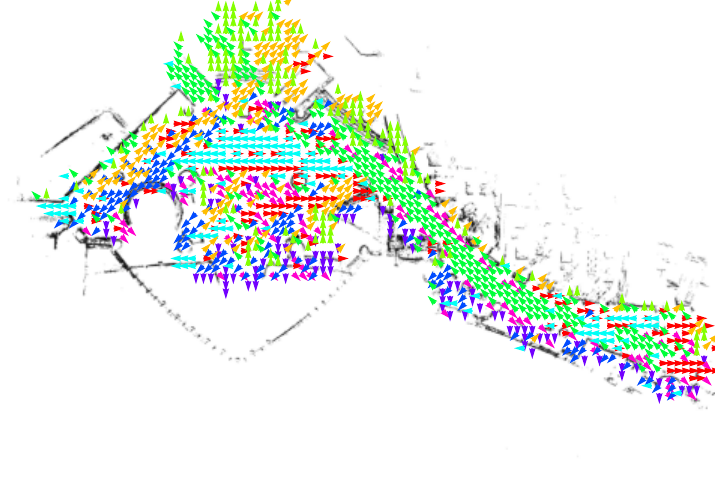}%
\includegraphics[clip,trim= 20mm 3mm 70mm 6mm,height=32mm]{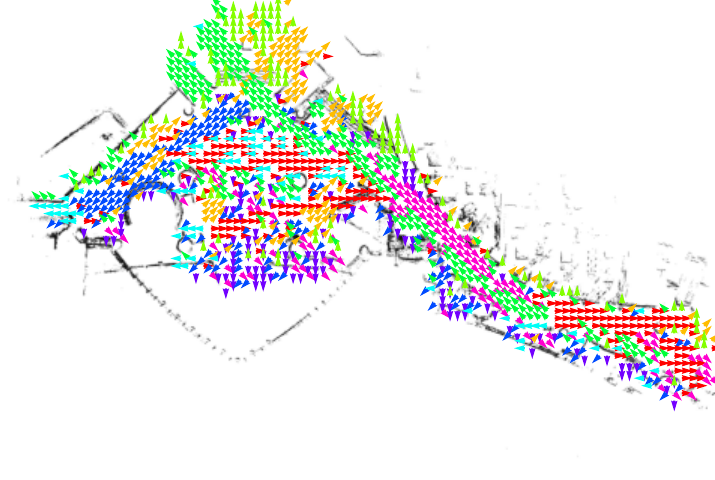}%
\includegraphics[clip,trim= 175mm 0mm 0mm 6mm,height=32mm]{figures/stef/stef-map-18.pdf}
\vspace*{-6mm}
\caption{STeF-map in the ATC dataset, for 10:00 (\textbf{left}), 14:00 (\textbf{middle}) and 18:00 (\textbf{right}), showing changes of motion patterns throughout the day represented by STeF-map. At each location, the colored arrow shows the dominant orientation for each cell in STeF-map,\add{where the arrow color encodes orientation.}}
\label{fig:stef-atc-time}
\vspace*{-7mm}
\end{figure*}

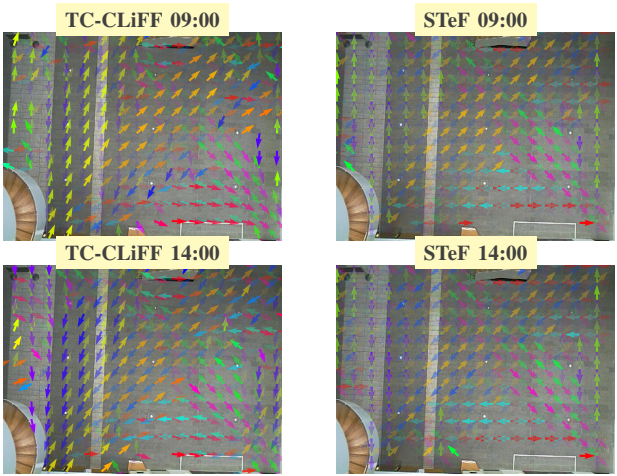
\begin{figure}
\centering
\begin{tikzpicture}
  % Upper left image
  \node[anchor=south west,inner sep=0] (image1) at (0,0) {\includegraphics[width=.43\linewidth]{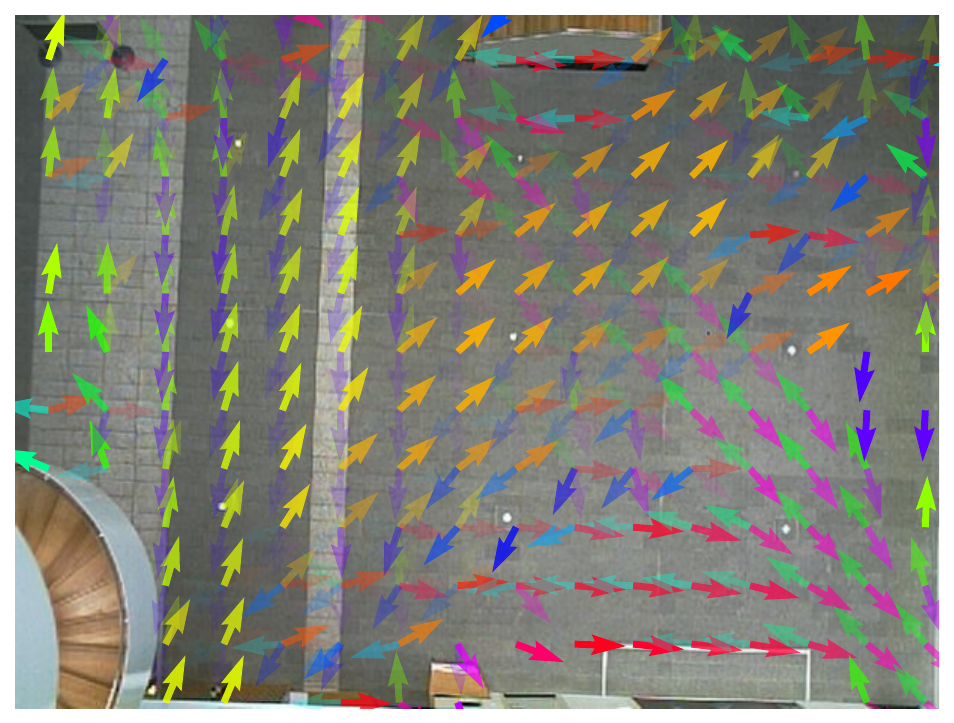}};
  \node[font=\bfseries\footnotesize, text=darkgray, fill=yellow!30] at  (image1.center) [xshift=0mm, yshift=15.5mm] {TC-CLiFF 09:00}; % Position text over image
  
  % Upper right image
  \node[anchor=south west,inner sep=0] (image2) at (0.5\linewidth,0) {\includegraphics[width=.43\linewidth]{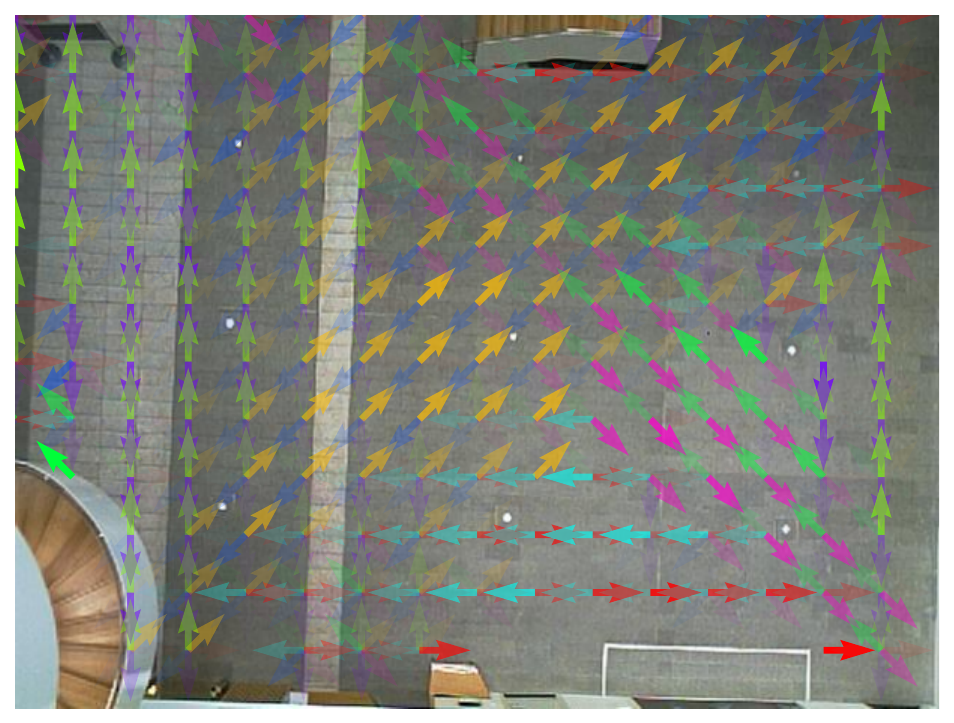}};
  \node[font=\bfseries\footnotesize, text=darkgray, fill=yellow!30] at (image2.center) [xshift=0mm, yshift=15.5mm] {STeF 09:00}; % Position text over image
  
  % Lower left image
  \node[anchor=south west,inner sep=0] (image3) at (0,-\linewidth*0.35) {\includegraphics[width=.43\linewidth]{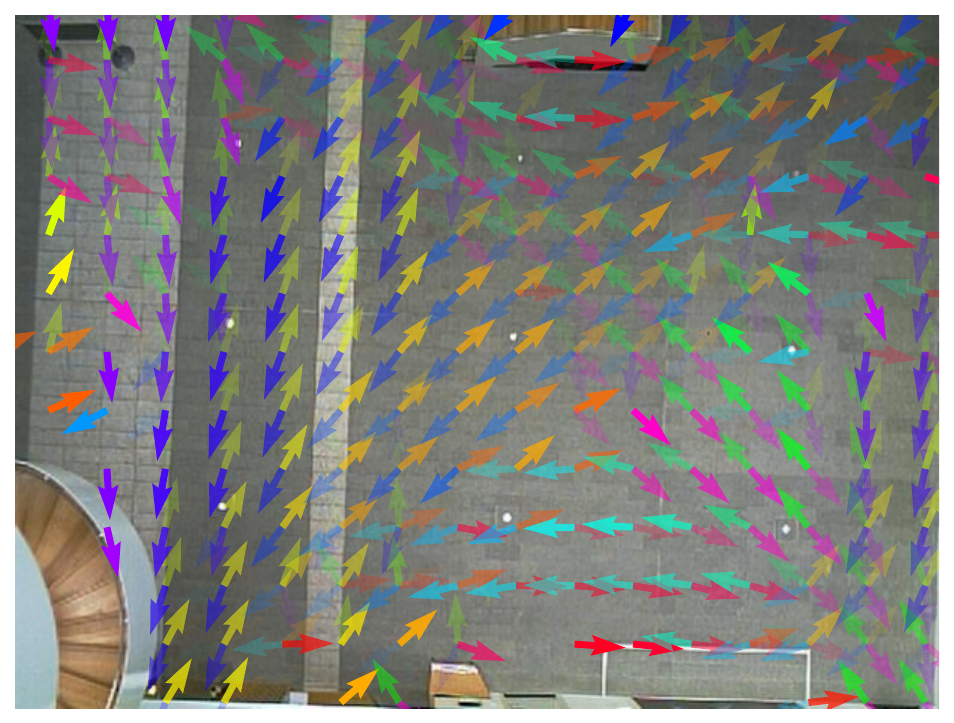}}; % Adjusted for spacing
  \node[font=\bfseries\footnotesize, text=darkgray, fill=yellow!30] at (image3.center) [xshift=0mm, yshift=15.5mm] {TC-CLiFF 14:00}; % Position text over image
  
  % Lower right image
  \node[anchor=south west,inner sep=0] (image4) at (0.5\linewidth,-\linewidth*0.35) {\includegraphics[width=.43\linewidth]{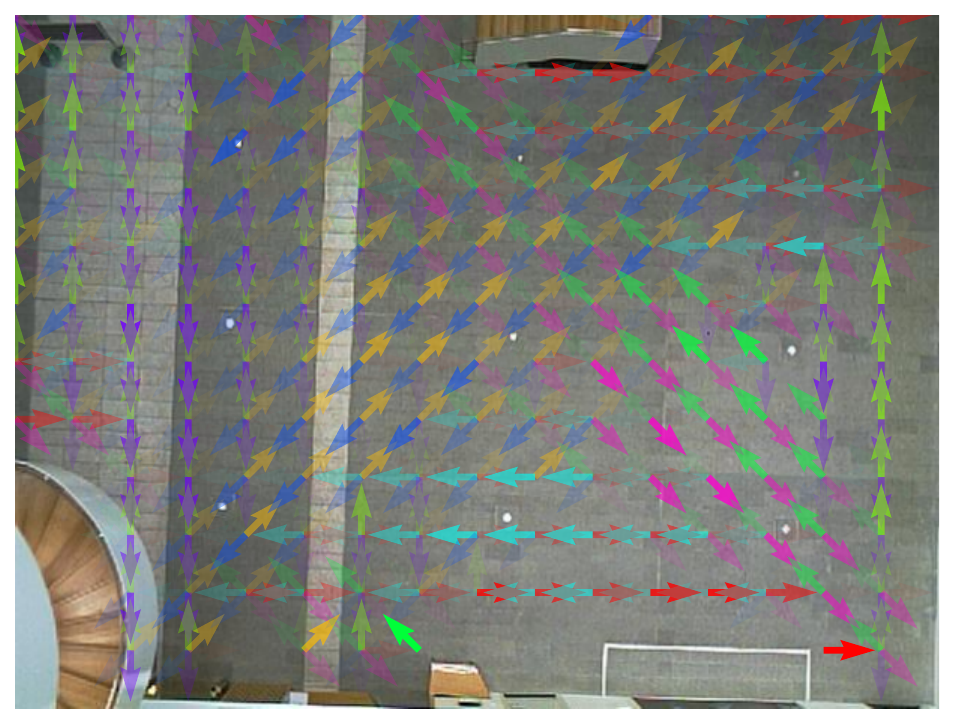}}; % Adjusted for spacing
  \node[font=\bfseries\footnotesize, text=darkgray, fill=yellow!30] at (image4.center) [xshift=0mm, yshift=15.5mm] {STeF 14:00}; % Position text over image
\end{tikzpicture}

\caption{Edinburgh dataset MoDs, showing motion pattern changes from 09:00 (\textbf{first row}) to 14:00 (\textbf{second row}). In Time-Conditioned CLiFF-map, colored arrows show the mean of the Gaussian component with transparency indicating the component weights. In STeF-map, colored arrows show each discretized orientation, with transparency reflecting their corresponding probabilities.\add{In both maps, arrow color encodes orientation.}}
\label{fig:edin_mod}
\vspace*{-4mm}
\end{figure}

\begin{figure}
\centering
\includegraphics[width=.42\linewidth]{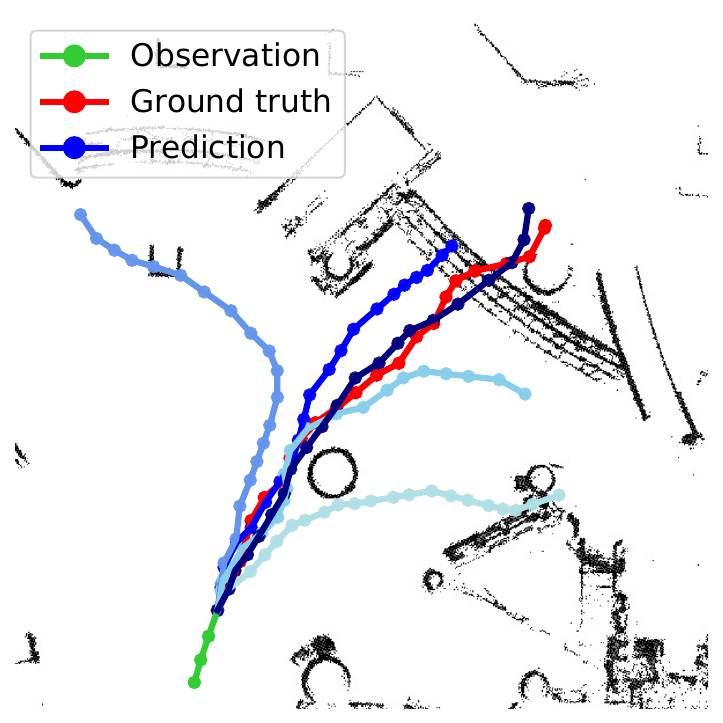}
\includegraphics[width=.42\linewidth]{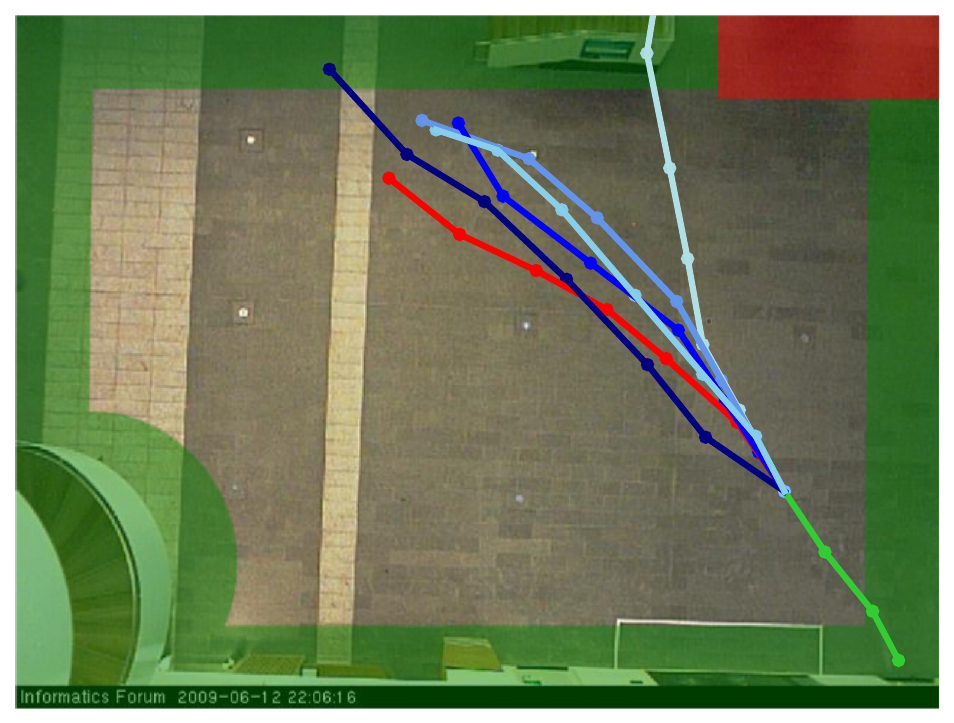}
\caption{Examples of predicted trajectory rankings using TC-CLiFF-LHMP in the ATC (\textbf{left}) and Edinburgh (\textbf{right}) datasets. The \textbf{red} line represents the ground truth trajectory, the \textbf{green} line represents the observed trajectory and \textbf{blue} lines the predicted trajectories, with darker shades of blue indicating higher-ranked predictions. Predictions in darker blue shows higher accuracy, showcasing the effectiveness of the ranking mechanism. \textbf{Right} figure also provides a view from the camera in the Edinburgh dataset. The green area shows the marginal region where trajectories start and end. The red area shows the region where people are waiting for a lift.}
\label{fig:rank}
\vspace*{-5mm}
\end{figure}

To predict human movement, the current time, $t_0$, determines the time interval.
%the future trajectory appears at.
The corresponding Time-Conditioned CLiFF-map is then used for prediction. The velocity sampling function remains the same as in the standard CLiFF-map and is illustrated in \cref{alg:samplecliff}.

\subsubsection{STeF-map} \label{method-stef}
STeF-map \cite{molina22exploration} is a spatio-temporal flow map, which models the likelihood of motion directions on a grid-based map by a set of harmonic functions. STeF-map captures long-term changes in crowd movements over time. Each grid cell contains $\kstef$ temporal models, corresponding to $\kstef$ discrete orientations of motion through the cell over time. 

%Each discretized orientation is denoted as $\thetastef$ ($\thetastef = i \frac{2\pi}{\kstef}$ and $i \in {0,1,...,\kstef-1}$).

The temporal models are based on the FreMEn framework \cite{krajnik2017fremen}, which uses the Fourier transform to model the probability of a given state as a function of time, represented by harmonic components. In the STeF-map model, the number of people detections in each orientation bin of each cell is counted in a predefined interval $t_\mathrm{stef}$. These counts are normalized, and the normalized values update the temporal model spectra. The frequency spectrum is analyzed, and the most $m_\mathrm{stef}$ prominent spectral components are transferred to the time domain.

\add{To predict motion pattern in a cell at a future time $t$, based on $m_\mathrm{stef}$ spectral components, STeF-map model computes a probability $p_{\theta_{\mathrm{stef}, i}}(t)$ for each discretized orientation $\theta_{\mathrm{stef}, i}$, where $\theta_{\mathrm{stef}, i} = i \frac{2\pi}{\kstef}$ and $i \in {0,1,...,\kstef-1}$, as described in~\cite{molina22exploration}. The final orientation assigned to the cell in STeF-map is the one with the highest predicted probability.}\cref{fig:stef-atc-time} shows the STeF-map at 10:00, 14:00 and 18:00 on the first day of the ATC dataset. In the Edinburgh dataset~\cite{majecka2009statistical}, the STeF-map of 09:00 (first row) and 14:00 (second row) are shown in \cref{fig:edin_mod}, compared with Time-Conditioned CLiFF-maps.\add{Implementation details are provided later in \cref{section-experiments}.}

%To use STeF-map for long-term human motion prediction, we sample the velocity from STeF-map, as in \cref{alg:samplestef}. With the given location ($x_t$, $y_t$), the corresponding cell in STeF-map is found.\add{While the standard STeF-map prediction selects the orientation with the highest predicted probability, we instead sample an orientation from the $\kstef$ discretized orientations, using their predicted probabilities as sampling weights.}Since STeF-map does not contain information of speed, in STeF-LHMP, we assume the speed is the same as in the last time step (line 3 of \cref{alg:samplestef}).

Since STeF-map encodes direction but not speed, STeF-LHMP assumes the speed remains constant, equal to that of the previous time step (Alg.~\ref{alg:samplestef}, line 3). The direction is sampled from the STeF-map cell at the query location ($x_t$, $y_t$).\add{While the standard STeF-map uses the direction with the highest predicted probability, we instead sample from $\kstef$ discretized orientations, using their predicted probabilities as weights.}

%STeF-map predicts human movement through a cell at a future time, by calculate the probability for each discretized orientation $\thetastef$ based on $m_\mathrm{stef}$ spectral components, as described in~\cite{molina22exploration}. 
%Instead of selecting the orientation with the highest probability (i.e., the dominant direction), we sample an orientation from discretized orientations $\thetastef$, according to the predicted probability distribution ${p_{\theta_i}(t)}_{i=0}^{\kstef-1}$.
\vspace{-2mm}
\subsection{Ranking predicted trajectories} \label{rank}
When evaluating MoD-LHMP, for an observed sequence, \cref{alg:LHMPAlgo} can be run multiple times to generate a number of predicted trajectories. Based on practical applications for autonomous robots, we rank the predicted trajectories and evaluate with the most likely output. 
The ranking is based on a fitness value associated with each predicted trajectory, derived from the sample velocities obtained from the MoD.
%To rank the output predicted trajectories, we use the probability of occurrence of the sample velocity obtained from the MoD.

To calculate the ranking of predicted trajectories, \cref{alg:LHMPAlgo} shows the updated part in red. At each prediction time step, when sampling a velocity from MoD, a corresponding fitness value is returned (line 7 of \cref{alg:LHMPAlgo}). For CLiFF-map and Time-Conditioned CLiFF-map, this fitness value corresponds to the likelihood of the sampled velocity under the Semi-Wrapped Gaussian Mixture Model (SWGMM). For STeF-map, the fitness value is given by the weight assigned to the selected discretized orientation. The overall fitness of a predicted sequence $\mathcal{T}$ is computed as the sum of per step fitness values in log space across $T_p$ prediction time steps.
%as the product of the fitness values of its sampled velocities across $T_p$ prediction time steps. 
An example of ranking is shown in \cref{fig:rank}.
\begin{table}[t]
\vspace{2mm}
    \centering
    %\captionsetup{skip=10pt}
    \begin{tabular}{lll}
     \toprule
        \textbf{Parameter}  & \textbf{ATC} & \textbf{Edinburgh} \\
        \midrule
        observation horizon $O_s$ & 3 \si{\second} & 3 \si{\second} \\
        max prediction horizon $T_s$ & 60 \si{\second} & 20 \si{\second} \\
        prediction time step $\Delta t$ & 1 \si{\second} & 1 \si{\second} \\
        CLiFF-map resolution & 1 \si{\metre} & 1 \si{\metre} \\
        STeF-map resolution & 1 \si{\metre} & 1 \si{\metre} \\
        sampling radius $r_s$ & 1 \si{\metre} & 1 \si{\metre} \\
        kernel parameter $\beta$ & 1 & 1 \\
        \bottomrule
    \end{tabular}
    \vspace{-1mm}
    \caption{Parameters used for ATC and Edinburgh datasets}
    \label{tab:scenarios}
\vspace*{-5mm}
\end{table}

%\begin{table*}[t]
%\vspace{2mm}
%    \centering
%    \begin{tabular}{lllllllll}
%     \toprule
%        \textbf{Dataset} & \textbf{$T_s$} & \multicolumn{7}{c}{\textbf{ADE/FDE} (m)} \\
%          & & TC-CLiFF & CLiFF & STeF & T++ & S-LSTM & LSTM & CVM \\
%        \midrule
%        ATC & 60 \si{\second} & \textbf{5.332} / \textbf{11.215} & 5.557 / 11.736 & 6.026 / 12.633 & 11.844 / 27.399 & 12.704 / %28.086 & 12.975 / 28.773 & 16.191 / 35.108 \\
%        Edinburgh & 20 \si{\second} & \textbf{3.035} / 5.787 & 3.094 / 5.830 & 3.064 / 5.847 & 3.810 / \textbf{4.897} & 3.835 / %5.195 & 3.835 / 5.201 & 7.919 / 17.112 \\
%        \bottomrule
%    \end{tabular}
%\vspace*{-1mm}
%    \caption{Long-term prediction horizon results in the ATC and Edinburgh datasets. $T_s$ is prediction horizon.}
%    \label{tab:expres}
%\vspace*{-5mm}
%\end{table*}

%% file: revision-content/experiments.tex
\section{Experiments} \label{section-experiments}
\subsection{Dataset}
%In this work, we introduce the MoD-LHMP class of methods, expanding upon the previously developed CLiFF-LHMP. We present two new methods: Time-Conditioned CLiFF-LHMP and STeF-LHMP, both designed to capture changes of human motion pattern over time. 
To evaluate MoD-LHMP with MoDs that capture changes of human motion patterns over time, it is essential to use datasets that span multiple days and reflect variations in human motion patterns throughout the day. Our experiments were conducted using two real-world datasets, ATC and Edinburgh, both of which provide sufficient multi-day coverage for evaluation. \add{Both datasets represent indoor open area.}

\subsubsection{ATC}
The ATC shopping center dataset~\cite{brscic2013person} contains real-world trajectories recorded in a shopping mall in Japan. This dataset covers a large indoor environment, with a total area covered of approximately \SI{900}{\unit{\metre\squared}}. %The map of the environment is shown in the left figure of \cref{fig:dataset_maps}. 
Given the extensive duration of the ATC dataset (92 days), we use a subset of 10 days in the experiments. The first day (Oct 24th) is used for training, and the remaining days are used for evaluation.

\subsubsection{Edinburgh}
The Edinburgh dataset \cite{majecka2009statistical} consists of pedestrian trajectories collected in the Informatics Forum building at the University of Edinburgh, covering around \SI{180}{\metre\squared} over several months of observation.
%Data were recorded on various days during this period, with the recording schedule not being consecutive. On some days, the recording sessions had gaps lasting for hours before recording resumed.  
Recordings were made on non-consecutive days, sometimes with hours-long gaps. To study the periodic patterns of human motion, we use all the dates when recordings were consecutive during daytime hours, from 08:00 to 16:00. In total, we use 45 days of data, with the first 5 for training and the remaining 40 for testing.
\add{For preprocessing, \textcite{majecka2009statistical} provides instructions on removing bad trajectories:} \begin{enumerate*} \item remove trajectories that start or end outside the marginal area of the scene (shown in green in the right figure of \cref{fig:rank}), \item remove trajectories shorter than 30 points because these could represent spurious detection, and \item remove trajectories that start and end in the area next to the lifts (shown in red in the right figure of \cref{fig:rank}), as these are likely produced by people waiting for a lift. \end{enumerate*}

%The covered environment is around \SI{180}{\unit{\metre\squared}}. %The map of the environment is shown in the right figure of \cref{fig:rank}. 

%The data is preprocessed according to instructions for removing bad trajectories \cite{majecka2009statistical}. We followed the same rule as stated in the Edinburgh dataset
%\begin{figure}[t]
%\centering
%\includegraphics[width=.59\linewidth]{figures/map/ATC-2.jpg}
%\includegraphics[width=.34\linewidth]{figures/map/EDIN.pdf}
% \vspace*{-2mm}
%\caption{\textbf{Left}: 2D map of ATC dataset. \textbf{Right}: The view from the camera in the Edinburgh dataset. The green area shows the marginal region where trajectories start and end. The red area shows the region where people are waiting for a lift.}
%\label{fig:dataset_maps}
%\vspace*{-3mm}
%\end{figure}

\subsection{Implementation Details}
Both the ATC and Edinburgh datasets in our experiments are downsampled to \SI{1}{\Hz}. For observations, we use \SI{3}{\second} of each trajectory and use the remaining (up to the maximum prediction horizon) as the prediction ground truth.

For the prediction horizon $T_s$, instead of using a fixed horizon, we explore a wider range of values in our experiments. We evaluate methods using prediction horizons up to a maximum value, which is determined based on the length distribution of each dataset. The $90^{th}$ percentile values are used as the maximum prediction horizons: \SI{60}{\second} for the ATC dataset and \SI{20}{\second} for the Edinburgh dataset.

To train CLiFF-maps and STeF-maps for both datasets, the map resolution is set to \SI{1}{\metre}. For STeF-map, we follow the parameters suggested in \cite{molina22exploration}. The number of discretized orientations, $\kstef$, is 8. The interval time, $t_\mathrm{stef}$, used for creating the STeF-map input histograms, is set to 10 minutes. The number of model components, $m_\mathrm{stef}$, is set to 2. %The resolution of STeF-map is \SI{1}{\metre}.

For CLiFF-LHMP and Time-Conditioned CLiFF-LHMP, the sampling radius $r_s$, which is used when selecting SWGMMs around the given location, is set to \SI{1}{\metre}. To bias the current orientation towards the sampled one, we use a default value of $\beta$ = 1 for both datasets. Prediction stops when no dynamics data (i.e. SWGMMs in CLiFF-map or temporal models in STeF-map) is available within the radius $r_s$ from the sampled location. For each ground truth trajectory, the prediction horizon $T_s$ is either equal to its length or to the maximum planning horizon for longer trajectories. The parameters used for both datasets are detailed in \cref{tab:scenarios}.

\add{\textbf{Privacy Considerations}: The datasets used for training are publicly available and fully anonymized, representing persons only as 2D positions without identifiers or visual data. MoDs further aggregate these trajectories into statistical motion patterns, so no %individual trajectories or 
personal information are retained.}

\subsection{Baseline}
We compare our method against Trajectron++~\cite{salzmann20}, LSTM-based prediction methods, a transformer-based model (Trajectory Unified TRansformer (TUTR)~\cite{shi2023tutr}), and a diffusion-based model (motion indeterminacy diffusion, MID)~\cite{gu2022mid}). 

Trajectron++ employs a graph-structured generative neural network based on a conditional-variational autoencoder. To implement Trajectron++ we used publicly available code and trained the model for 100 epochs on the training data of both datasets. %Parameter configurations are provided with project code. 
Social LSTM~\cite{alahi2016social} and vanilla LSTM are used as comparison baseline representing LSTM-based methods. TrajNet++ framework~\cite{Kothari2020HumanTF} is used for implementing LSTM and S-LSTM. For training LSTM-based models, we used the parameters proposed in the Social LSTM method, with the pooling size set at 32, sum pooling window size set to be 8 $\times$ 8, the dimension of hidden state for all LSTM models at 128, and the learning rate set at 0.003.

\add{TUTR unifies social interaction modeling and multimodal trajectory prediction components in a transformer encoder-decoder architecture. It predicts multiple trajectories with corresponding probabilities, and we evaluate using the trajectory with the highest probability. MID formulate trajectory prediction as a reverse process of motion indeterminacy diffusion, which gradually discards the indeterminacy to obtain desired trajectory from ambiguous walkable areas. Both MID and TUTR are trained for 100 epochs on both datasets.}

\vspace{-2mm}
\subsection{Evaluation Metrics}
For the evaluation of predictive performance we used the following metrics: \emph{Average} and \emph{Final Displacement Errors} (ADE and FDE). ADE describes the mean $L^2$ distance between predicted trajectories and the ground truth. FDE describes the $L^2$ distance between the predicted and the ground truth positions at the last prediction time step. For each ground truth trajectory we generated $k$~=~5 prediction trajectories and evaluate with the most likely output with the proposed ranking method.\add{When probability information for the predicted trajectories is not available, we report the mean ADE and FDE values over the predicted trajectories. This applies to the experiments on the no-ranking method (results shown in \cref{fig:rank-metrics}) and to the baseline method, MID, since MID does not provide probability information for its predictions.}

%% file: revision-content/results.tex
\section{Results} \label{section-result}
% In this section, we present the results obtained from the ATC and Edinburgh datasets using the MoD-LHMP method and various baselines. The performance is evaluated with both quantitative and qualitative analysis.

\begin{table}[t]
    \centering
    \begin{tabular}{lll}
     \toprule
        \textbf{Method} & \multicolumn{2}{c}{\textbf{ADE/FDE} (m)} \\
        & ATC ($T_s=60$) & Edinburgh ($T_s=20$) \\
        \midrule
        TC-CLiFF & \textbf{5.332} / \textbf{11.215} & \textbf{3.035} / 5.787 \\
        CLiFF & 5.557 / 11.736 & 3.094 / 5.830 \\
        STeF & 6.026 / 12.633 & 3.064 / 5.847 \\
        T++ & 11.844 / 27.399 & 3.810 / 4.897 \\
        MID & 21.124 / 44.373 & 5.621 / 7.031 \\
        TUTR & 12.127 / 26.739 & 3.375 / \textbf{4.432} \\
        S-LSTM & 12.704 / 28.086 & 3.835 / 5.195 \\
        LSTM & 12.975 / 28.773 & 3.835 / 5.201 \\
        CVM & 16.191 / 35.108 & 7.919 / 17.112 \\
        \bottomrule
\end{tabular}
\label{tab:expres}
\caption{\add{Long-term prediction horizon results in the ATC and Edinburgh datasets. $T_s$ is prediction horizon.}}
\vspace*{-2mm}
\end{table}

\begin{figure}[t]
\centering
\includegraphics[width=.48\linewidth]{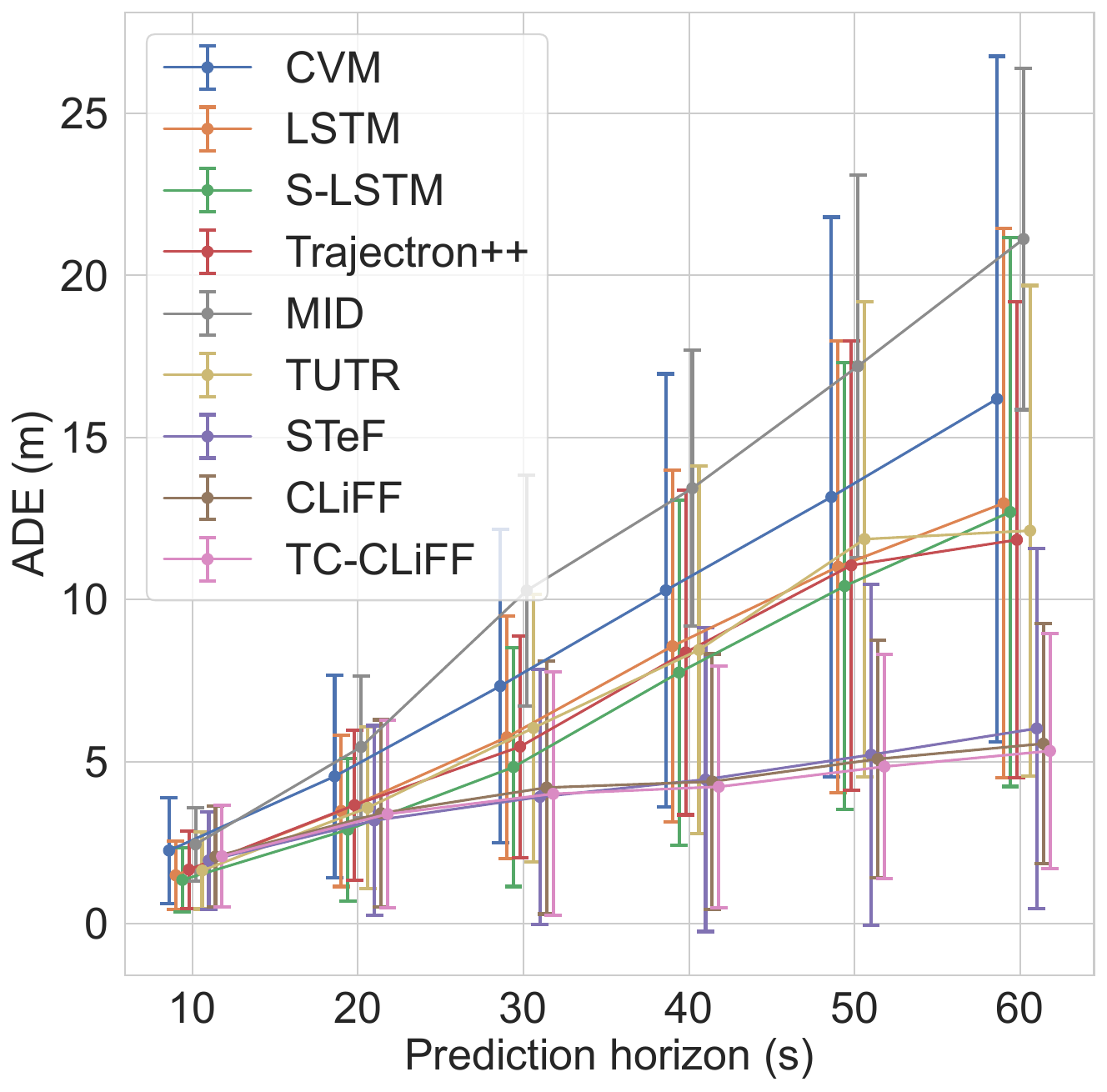}
\includegraphics[width=.48\linewidth]{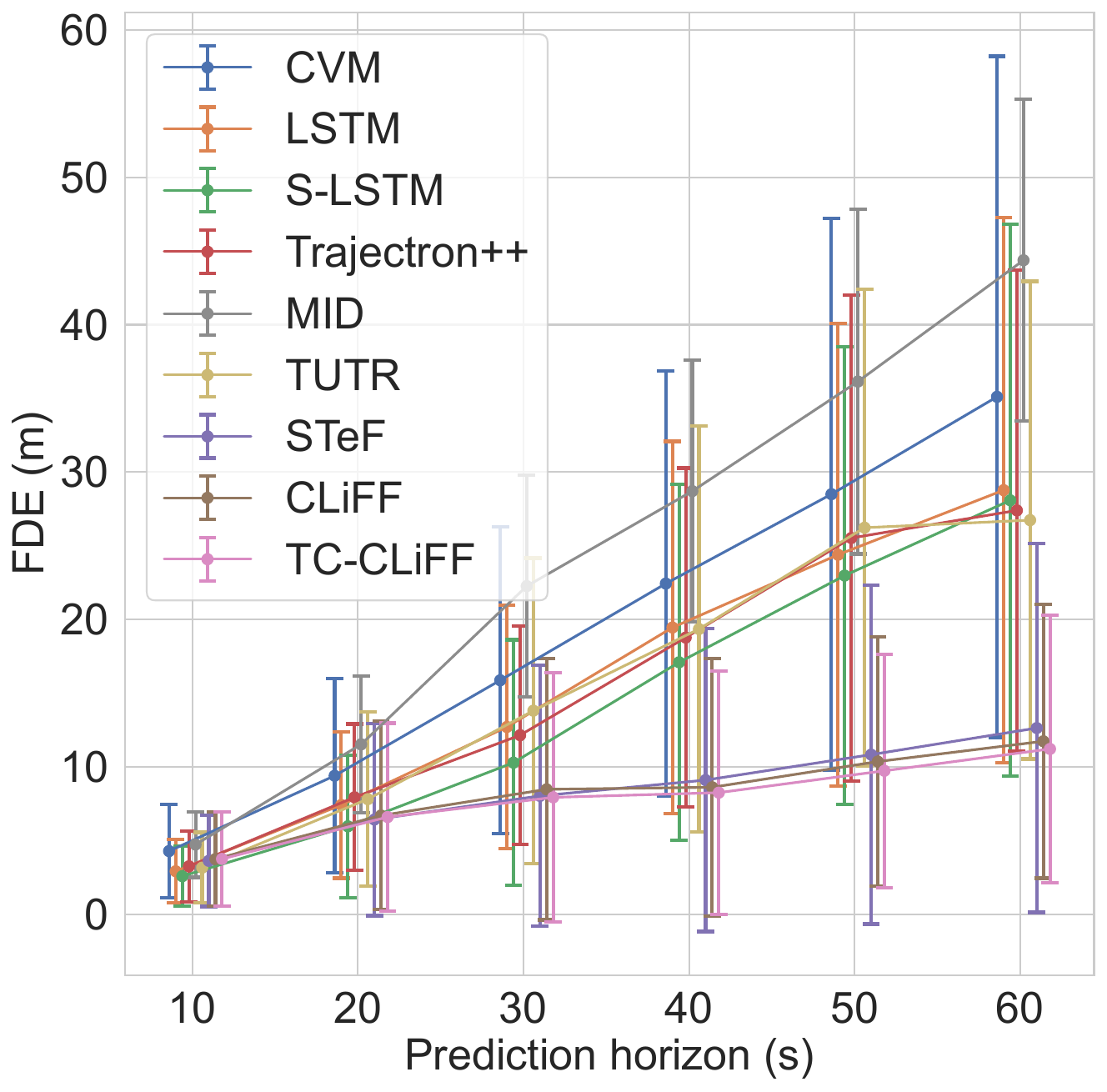}
%\vspace*{-2mm}
\caption{\add{ADE/FDE (mean $\pm$ one std. dev.) in the ATC dataset with planning horizon 1--\SI{60}{\second}.}}
\label{fig:atc_final}
\vspace*{-4mm}
\end{figure}

\begin{figure}[t]
\centering
\includegraphics[width=.48\linewidth]{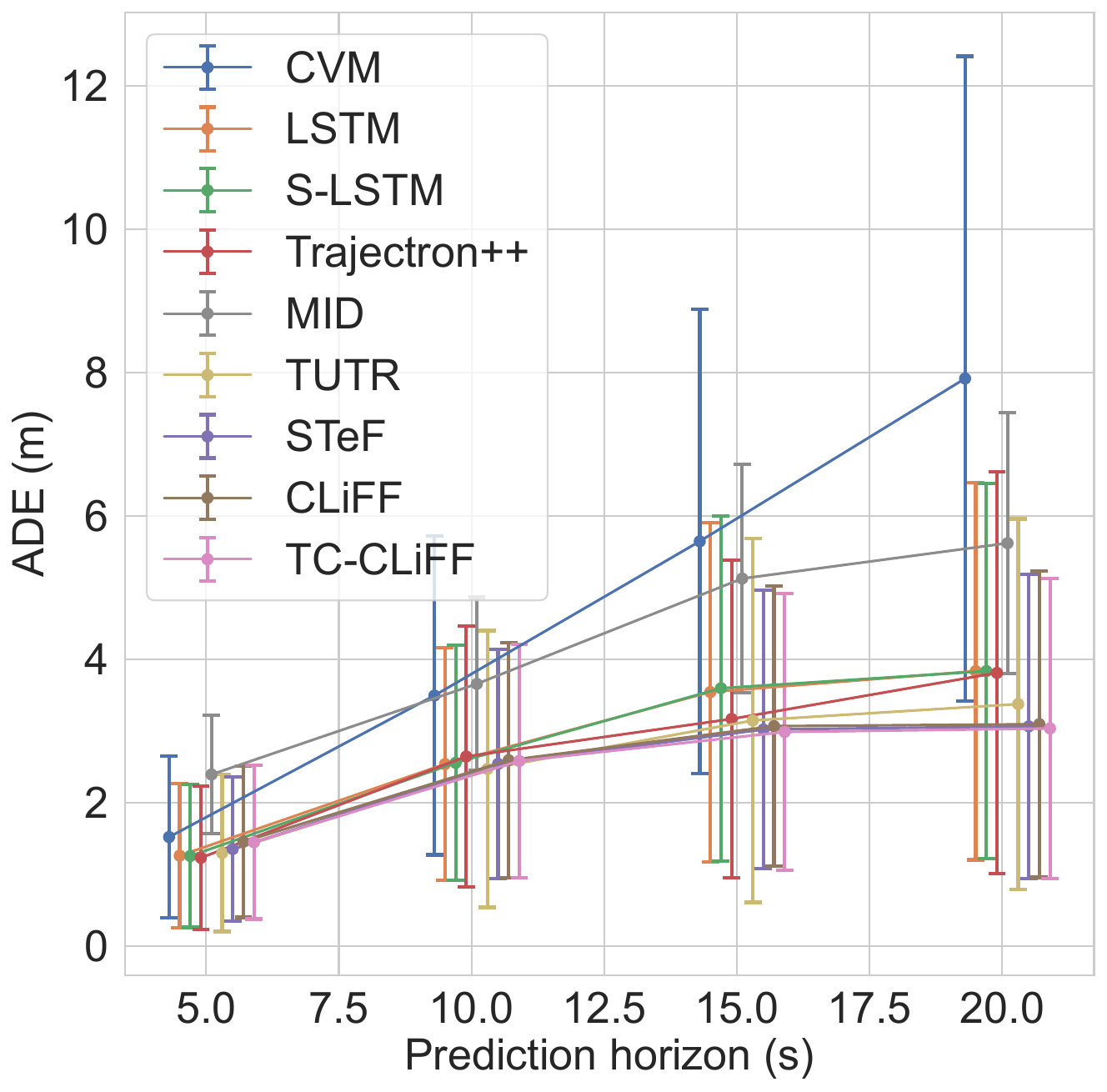}
\includegraphics[width=.48\linewidth]{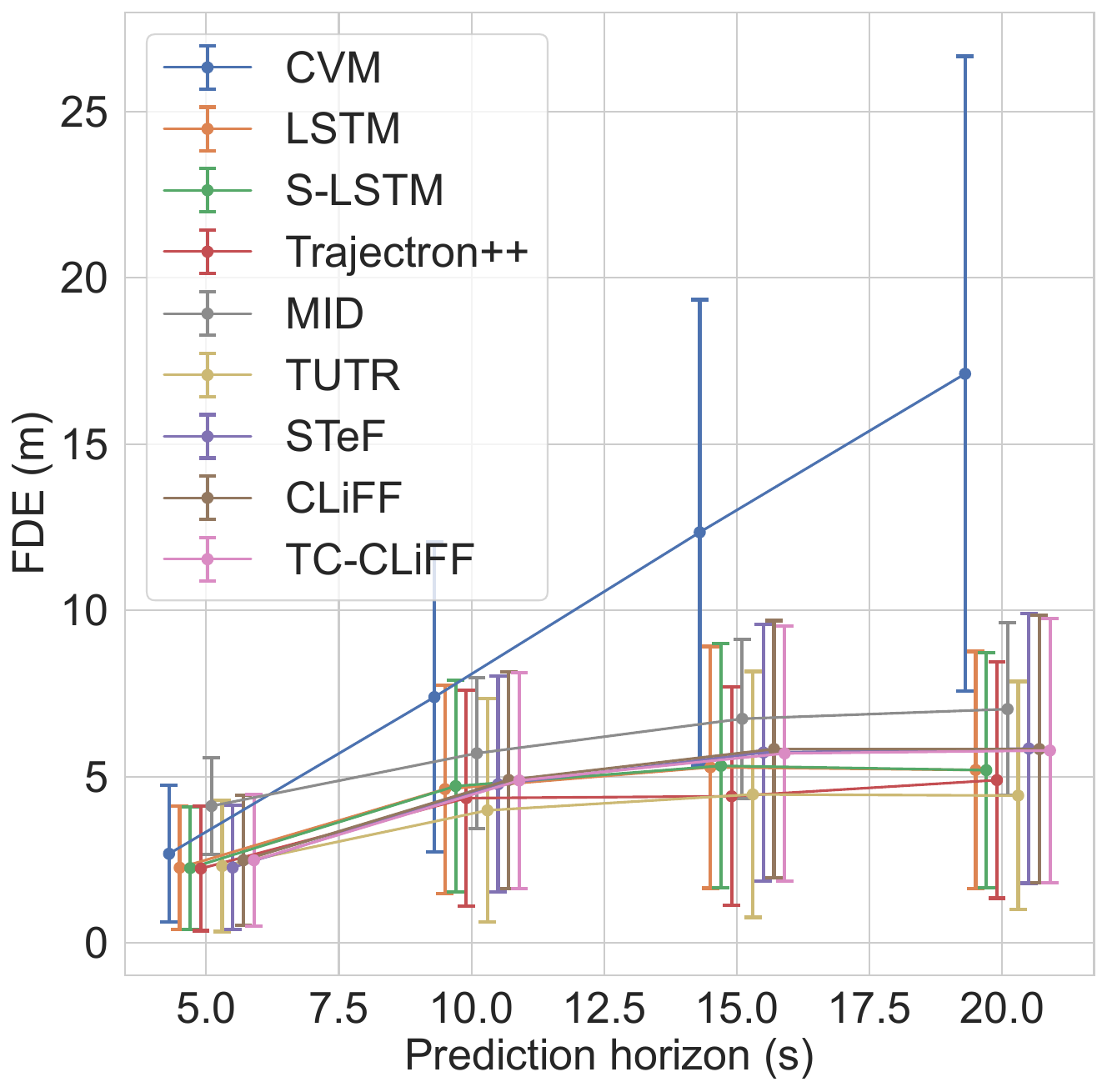}
%\vspace*{-2mm}
\caption{\add{ADE/FDE (mean $\pm$ one std. dev.) in the Edinburgh dataset with planning horizon 1--\SI{20}{\second}.}}
\label{fig:edin_final}
\vspace*{-3mm}
\end{figure}

\begin{figure}[t]
\centering
\includegraphics[width=.45\linewidth]{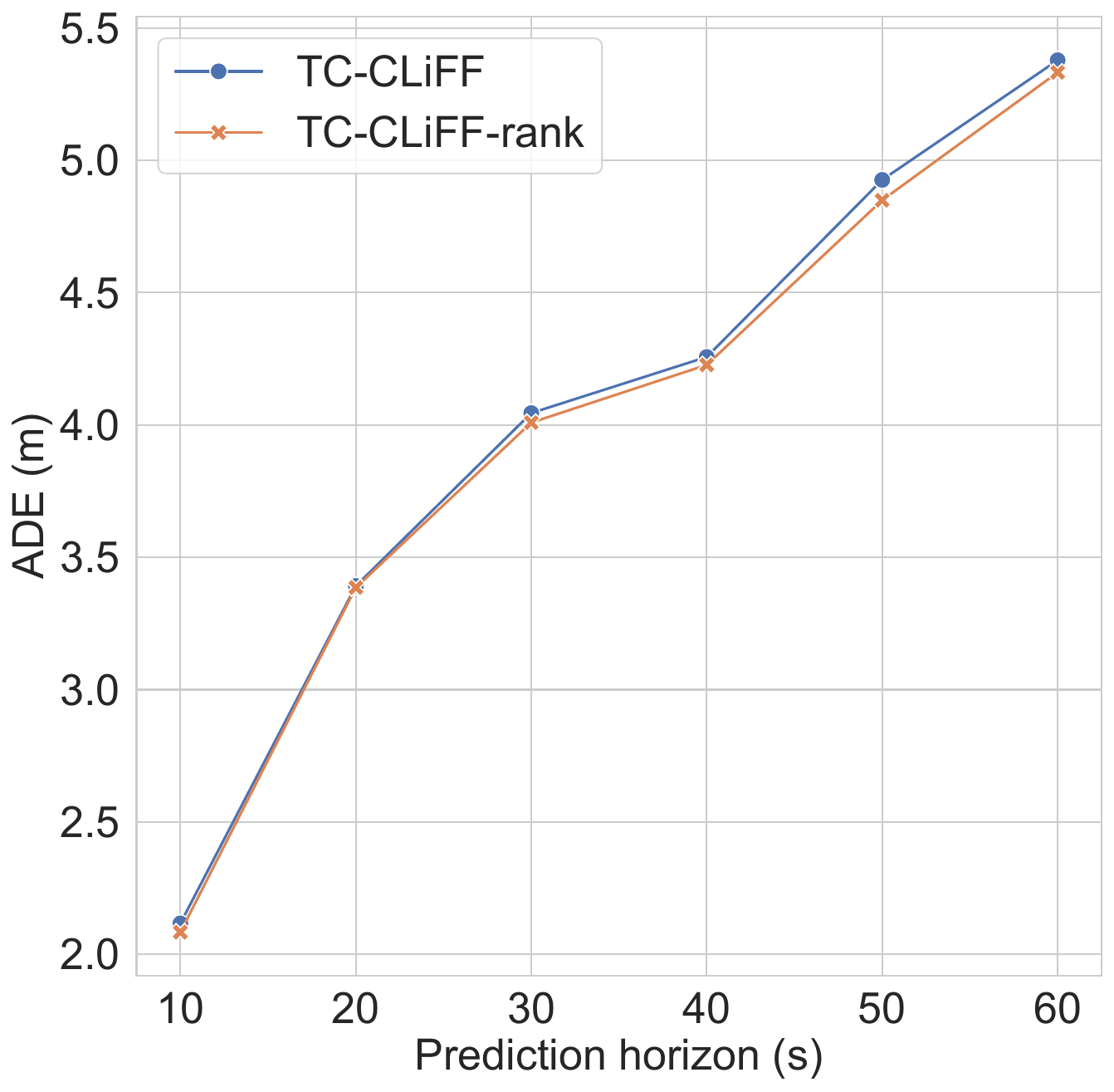}
\includegraphics[width=.45\linewidth]{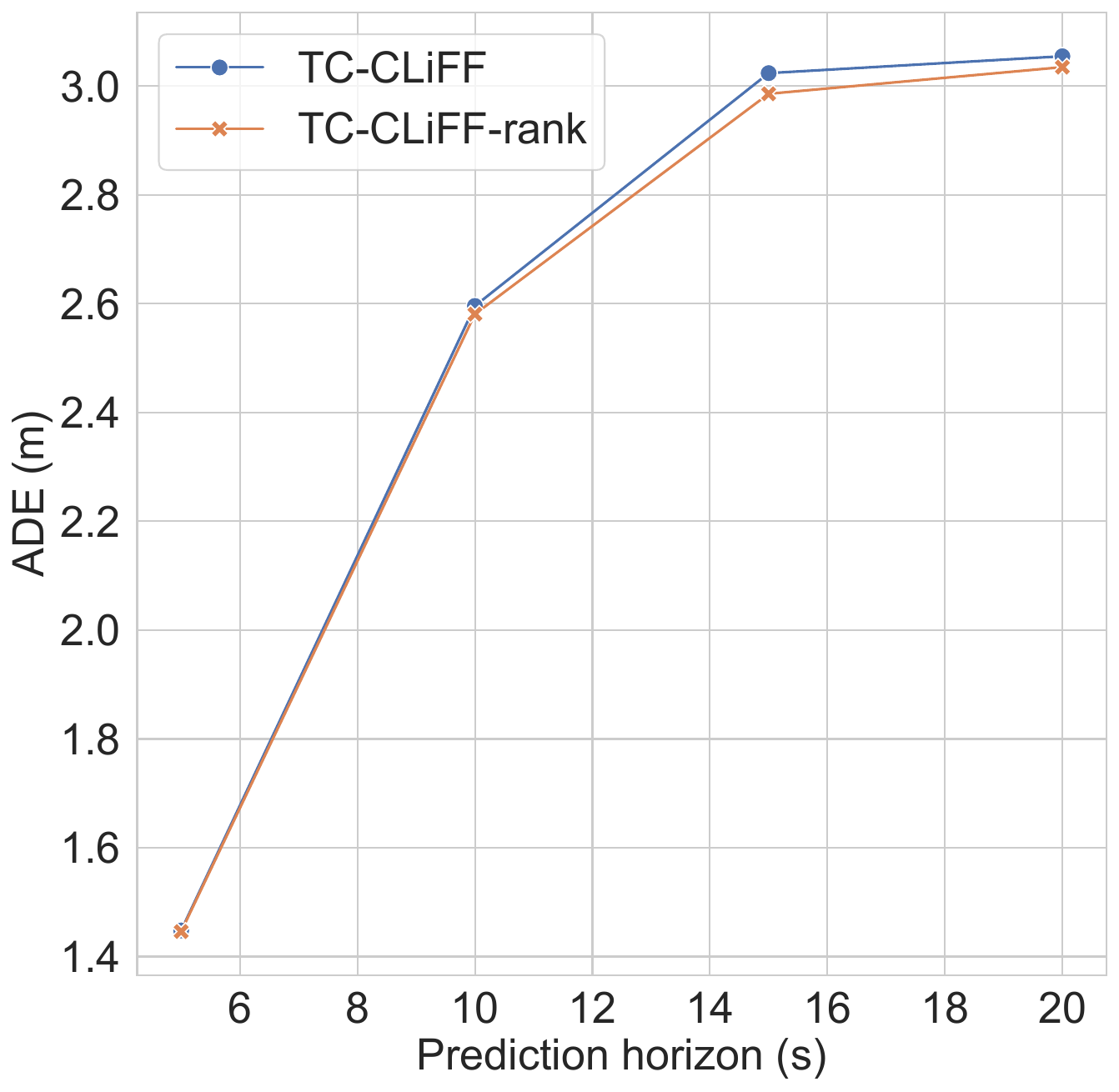}
%\vspace*{-2mm}
\caption{ADE of Time-Conditioned CLiFF-LHMP in the ATC (\textbf{left}) and Edinburgh (\textbf{right}) dataset with and without ranking.}
\label{fig:rank-metrics}
\vspace*{-3mm}
\end{figure}

\begin{figure}[t]
\centering
\includegraphics[width=.48\linewidth]{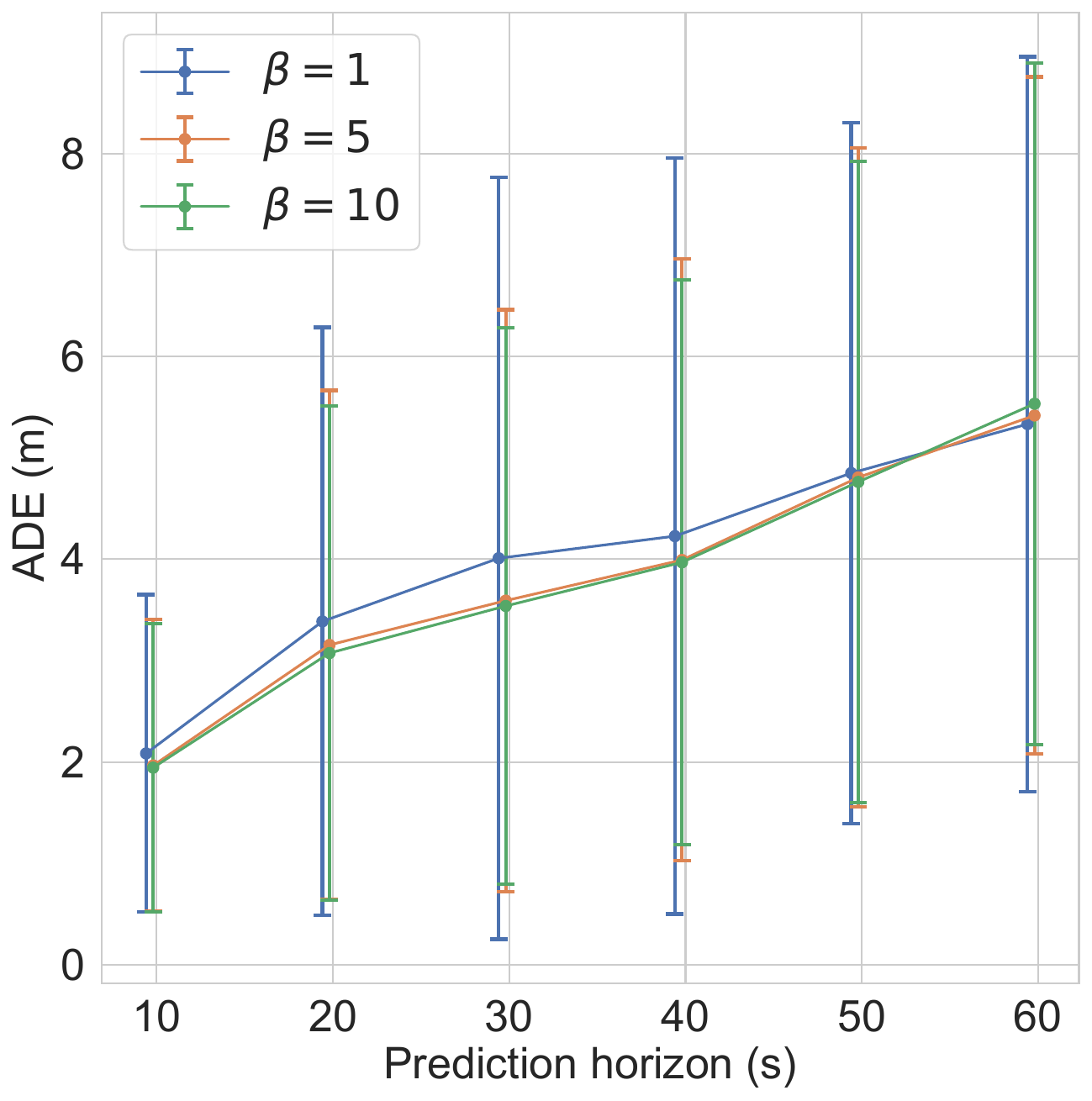}
\includegraphics[width=.48\linewidth]{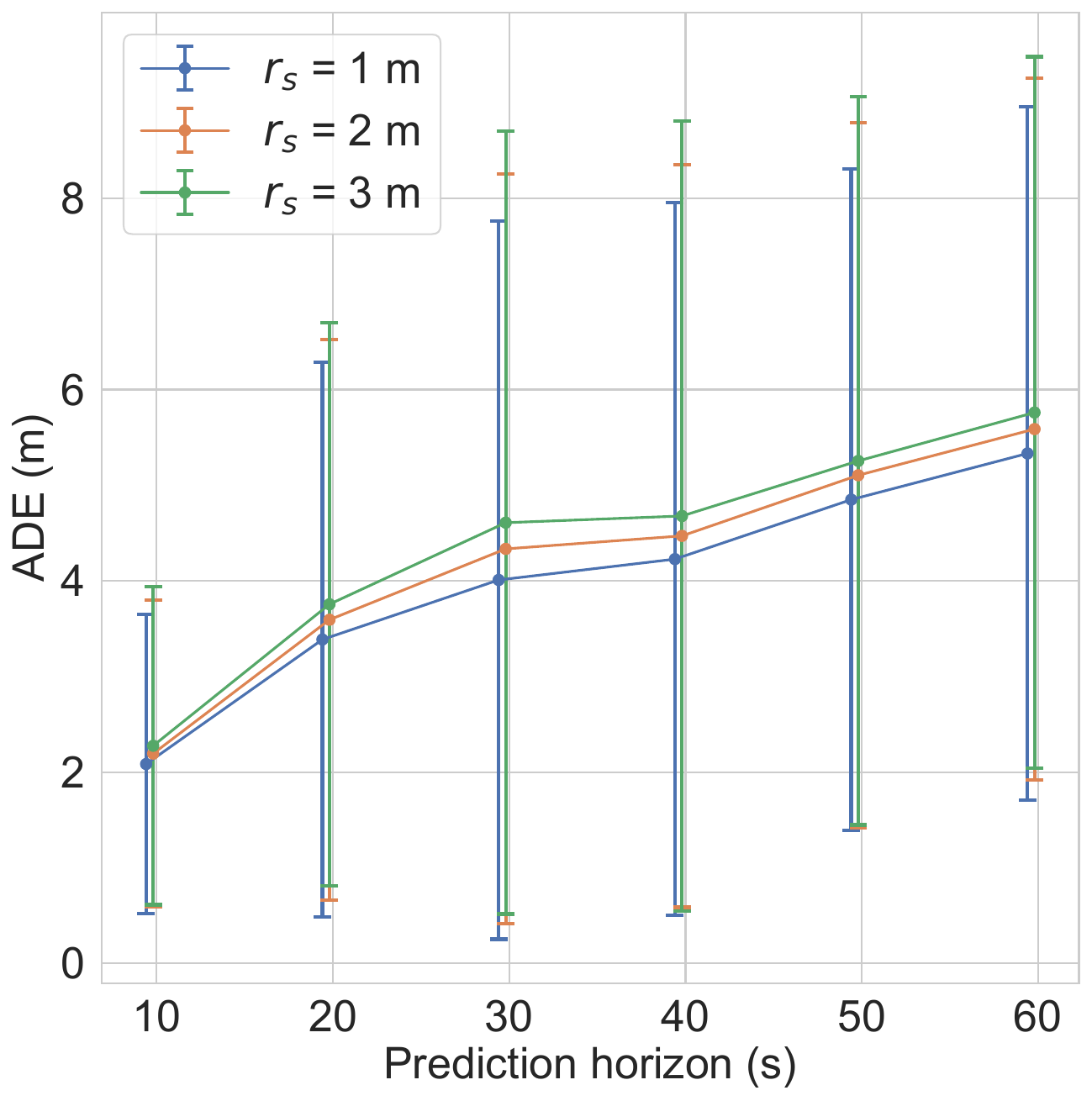}
%\vspace*{-2mm}
\caption{Parameter analysis on the ATC dataset, for TC-CLiFF-LHMP method, showing the ADE (mean $\pm$ one std. dev.) over different prediction horizons vs the kernel parameter $\beta$ \textbf{(left)} and sampling radius $r_s$ \textbf{(right)}.}
\label{fig:parameter}
\vspace*{-3mm}
\end{figure}

\begin{figure}[t]
\centering
\includegraphics[clip,trim= 0mm 0mm 0mm 0mm,height=30mm]{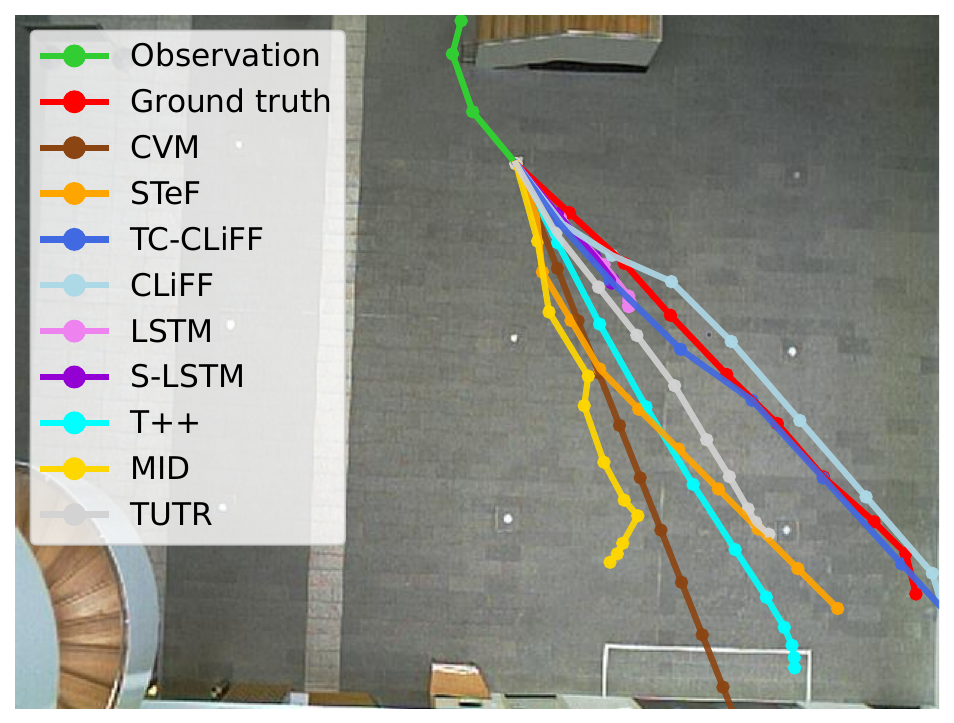}
\includegraphics[clip,trim= 35mm 20mm 60mm 0mm,height=30mm]{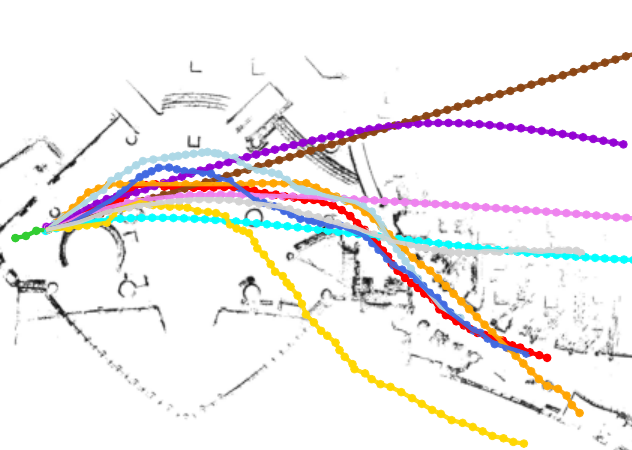}
%\vspace*{-2mm}
\caption{Prediction examples in Edinburgh dataset (\textbf{left}) with a \SI{10}{\second} prediction horizon and the ATC dataset (\textbf{right}) with a \SI{60}{\second} horizon. %The \textbf{red} line represents the ground truth trajectory and the \textbf{green} line represents the observed trajectory. 
MoD-LHMP methods make more accurate predictions than deep learning methods. In the ATC dataset, when the trajectory predicted by S-LSTM, LSTM, Trajectron++,\add{TUTR and MID}are unfeasible by crossing the walls, MoD-LHMP predictions are following the topology of the environment.}
\label{fig:prediction_example}
\vspace*{-4mm}
\end{figure}

\subsection{Quantitative evaluation}
\cref{tab:expres} presents the evaluation results for the ATC and Edinburgh datasets, featuring MoD-LHMP methods, which includes Time-Conditioned CLiFF-LHMP (TC-CLiFF), CLiFF-LHMP (CLiFF), STeF-LHMP (STeF), as well as deep learning-based methods including\add{Trajectory Unified TRansformer (TUTR), motion indeterminacy diffusion (MID)}, Trajectron++ (T++), Social LSTM (S-LSTM) and vanilla LSTM. Constant velocity model (CVM) is also compared as the baseline. The table shows the performance at the maximum prediction horizon for both datasets. Results for other prediction horizon are detailed in \cref{fig:atc_final} for the ATC dataset and \cref{fig:edin_final} for the Edinburgh dataset. 

In short-term predictions, at \SI{10}{\second} for ATC and \SI{5}{\second} for Edinburgh, deep learning-based methods perform slightly better than the MoD-based ones. However, as the prediction horizon increases, MoD-based methods achieve better accuracy. At the prediction horizon \SI{60}{\second}, TC-CLiFF-LHMP achieves over 50\% better ADE accuracy than the deep learning-based methods,\add{with paired t-tests showing $p<0.001$, under the null hypothesis that our method yields equal or higher mean error than the baseline.}In the Edinburgh dataset, whose environment area is about 20\% of ATC, deep learning-based methods achieve lower FDE. This is due to idling behaviors in Edinburgh, where pedestrians stop or slow down at the end of a path. Sequence-based models such as LSTMs and transformers capture this pattern and generate slowed predictions, reducing FDE. MoD-based methods can also capture idling if it frequently occurs at the same location (e.g., a resting point), but when the dominant motion pattern corresponds to normal walking speed, they have a higher probability of sampling continued motion rather than idling, thereby increasing FDE in such trajectories. In contrast, the ATC dataset spans a much larger area with longer, more continuous trajectories, where local motion patterns captured by CLiFF-maps provide a stronger advantage for long-horizon prediction, resulting in a larger performance margin over the baselines.

%\add{To further assess data requirements, we expanded the ATC training window from a single day (Oct 24, 2012) to 10 days (between Nov 28, 2012 and Jan 6, 2013 for MID and TUTR baselines. Since a single day of the ATC dataset already provides large amount of data, extending the training window to 10 days yields only limited changes for the baseline methods (MID: 21.124/44.373 to 19.767/41.385, TUTR:12.127/26.739), confirming that the advantage of MoD-based methods is not limited to single-day training.}
%In the Edinburgh dataset, whose environment area is approximately 20\% of ATC dataset, deep learning-based methods predict goal locations more accurately at a \SI{20}{\second} prediction horizon, achieving better FDE results, although the average displacement error over the predicted trajectory (ADE) is larger. 

Within the MoD-LHMP framework, TC-CLiFF-LHMP consistently outperforms the standard CLiFF-LHMP,\add{with paired t-tests showing $p<0.001$ for both ADE and FDE, under the null hypothesis that our method yields equal or higher mean error than the baseline.}Given that motion patterns change over time in both datasets, TC-CLiFF-LHMP captures shifts in them and achieves better accuracy. This improvement remains at the higher prediction horizons. In the long-term, particularly over \SI{30}{\second} in the ATC dataset, CLiFF-based methods outperform STeF-LHMP,\add{with paired t-tests showing $p<0.001$ for both ADE and FDE, under the same null hypothesis.}Unlike STeF-map which uses a discrete 8-bin histogram of orientation distribution and ignores speed, CLiFF-map models a continuous joint distribution of speed and orientation, achieving more accurate long-term predictions.

The evaluation of the ranking method in CLiFF-LHMP, which outputs the most likely predicted trajectory, is shown in \cref{fig:rank-metrics}. Its benefits increase continuously with the prediction horizon, becoming more pronounced as it extends.

\textbf{Runtime analysis}: 
\add{At each prediction step, MoD-LHMP queries the MoD, with $M$ stored locations, to find location neighbors and samples a velocity, followed by a constant-time kernel update. The total complexity for $N$ trajectories over a horizon of $T_p$ time steps is $\mathcal{O}(N T_p M)$, which scales linearly with both horizon and dataset size. For inference performance measurements, we used a laptop running Ubuntu 20.04 with an Intel Core i7-1185G7 CPU. When evaluating TC-CLiFF-LHMP, for ATC dataset, the average inference time is \SI{0.11}{\second} per trajectory for prediction horizon up to \SI{60}{\second}. For Edinburgh dataset, the average inference time is \SI{0.04}{\second} per trajectory for prediction horizon up to \SI{20}{\second}. 
CPU-only inference supports $\sim$\SI{10}{\hertz} trajectory prediction rates, which are sufficient for embedded deployment.
The inference process maintains a steady resident memory footprint of approximately 273 MB on the laptop CPU. The storage size of each Time-Conditioned CLiFF-map is about 152 kB per hour on average.}

\subsection{Parameter Analysis}
\Cref{fig:parameter} presents a sensitivity analysis for the parameters of CLiFF-based methods: kernel parameter $\beta$ and sample radius $r_s$. 
The parameter $\beta$ scales CLiFF-map term based on the difference between the sampled and current directions. A higher $\beta$ value makes CLiFF-LHMP to behave more like a CVM, whereas a lower $\beta$ value results in predictions that more closely follow the sampled velocity. As the prediction horizon increases, a lower $\beta$ value achieves better performance, showing that relying more on the CLiFF-map enhances accuracy. The sample radius $r_s$ determines the selected nearby SWGMMs within the CLiFF-map. 
With a sample radius $r_s$ close to the map resolution (\SI{1}{\metre}), more accurate motion patterns can be captured, leading to better prediction accuracy.

\vspace{-3mm}
\subsection{Qualitative evaluation}
\cref{fig:prediction_example} presents qualitative prediction examples. In both datasets, MoD-LHMP outperform deep learning methods. In the ATC environment, as no explicit knowledge of the obstacle layout is provided, deep learning methods predict infeasible trajectories, such as crossing walls. In contrast, MoD-based methods leverage learned motion patterns in MoDs to predict realistic trajectories that follow the complex topology of the environment, implicitly taking obstacles into account.

%% file: revision-content/conclusions.tex
\section{Conclusion} \label{section-conclusions}
In this work, we introduce MoD-informed LHMP framework, which is compatible with various types of MoDs. To handle dynamics human flow variations, we present a Time-Conditioned CLiFF-LHMP, which adapts to changing motion patterns throughout the day. Experiments on two real-world datasets show that MoD-informed LHMP approaches outperform state-of-the-art deep learning methods, and that incorporating temporal information further improves prediction accuracy.
As the current method discretizes both spatial and temporal dimension into uniform grids, future work will include continuous mapping of human dynamics in spatial and temporal domains, enhancing representation accuracy.\add{To further improve prediction performance, learning the kernel parameter $\beta$ online from residual errors could help the model adapt to scene-specific dynamics.}